\documentclass[10pt,twocolumn,letterpaper]{article}

\usepackage[pagenumbers]{cvpr} 

\definecolor{cvprblue}{rgb}{0.21,0.49,0.74}
\usepackage[pagebackref,breaklinks,colorlinks,allcolors=cvprblue]{hyperref}
\usepackage{makecell}
\usepackage{booktabs}
\usepackage{colortbl}
\usepackage{xcolor}
\usepackage{makecell}
\usepackage{booktabs}
\usepackage{makecell}
\usepackage[table]{xcolor}
\usepackage{colortbl}
\usepackage[normalem]{ulem}
\usepackage{caption}
\usepackage{array}
\usepackage{multicol}

\usepackage{float}
\usepackage[normalem]{ulem} 
\usepackage{tcolorbox}
\tcbuselibrary{breakable}

\usepackage[table]{xcolor}
\usepackage{amssymb}
\usepackage{pifont}
\newcommand{\cmark}{\ding{51}}%
\newcommand{\xmark}{\ding{55}}%
\usepackage[normalem]{ulem}
\definecolor{headergray}{RGB}{245,245,245}
\definecolor{groupgray}{RGB}{250,250,250}
\definecolor{ourscache}{HTML}{E8F5E9}    
\definecolor{oursnocache}{HTML}{E3F2FD}  
\usepackage{xcolor}  
\def\paperID{3920} 
\def\confName{CVPR}
\def\confYear{2026}

\begin{document}

\title{FreeSwim: Revisiting Sliding-Window Attention Mechanisms for\\Training-Free Ultra-High-Resolution Video Generation}


\author{Yunfeng Wu$^{*12}$, Jiayi Song$^{*13}$, Zhenxiong Tan$^{4}$, Zihao He$^{1}$, and Songhua Liu$^{\dagger1}$\\
$^{1}$School of Artificial Intelligence, Shanghai Jiao Tong University,\\
$^{2}$Xi'an Jiaotong-Liverpool University, $^{3}$Xi'an Jiaotong University, $^{4}$National University of Singapore
}

\twocolumn[{
\renewcommand\twocolumn[1][]{#1}
\maketitle
\begin{center}
\vspace{-0.3cm}
    \captionsetup{type=figure}
    \includegraphics[width=\textwidth]{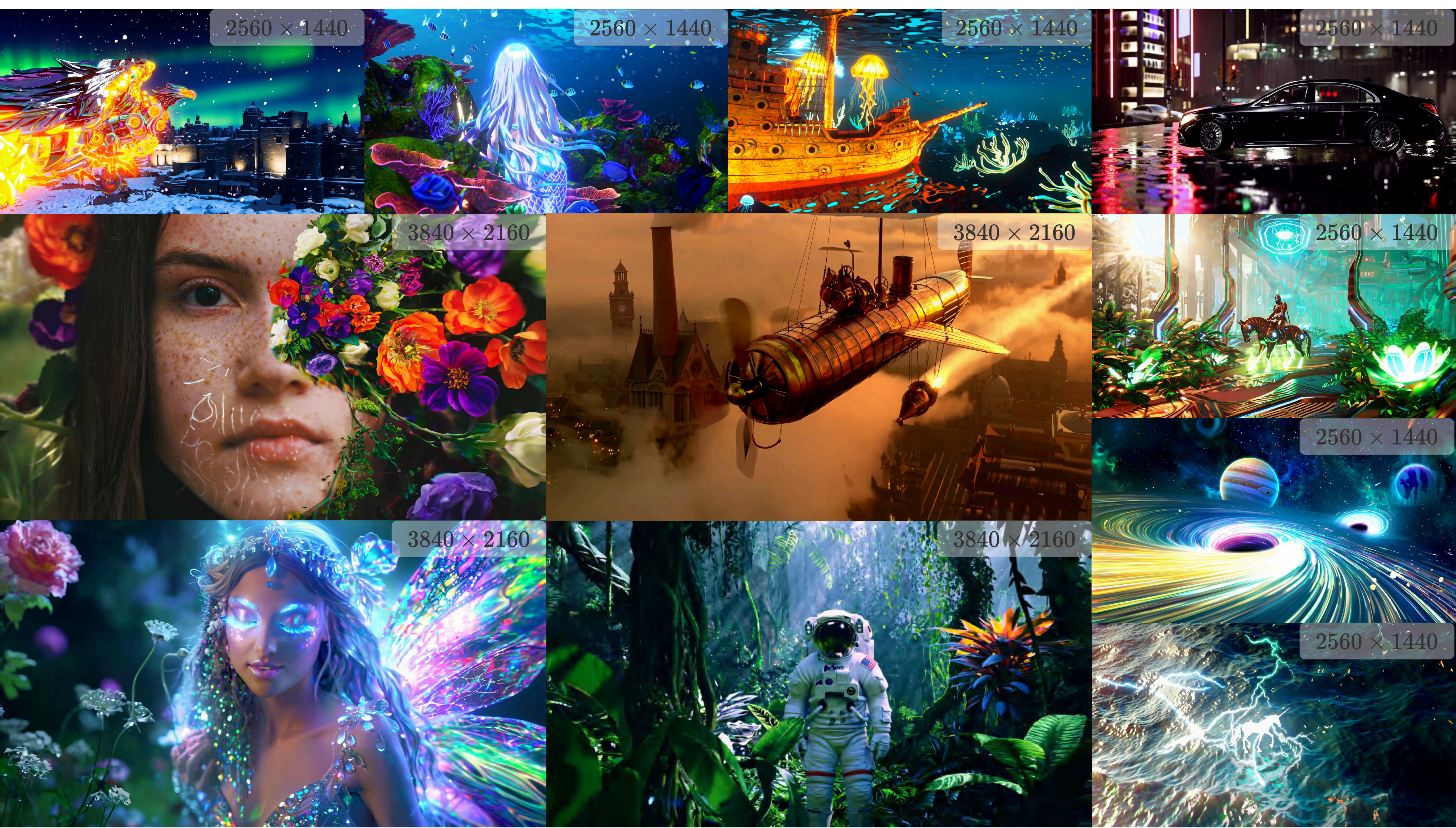}
    \vspace{-0.7cm}
    \captionof{figure}{Ultra-resolution results generated by our FreeSwim built upon Wan2.1~\cite{wan2025wanopenadvancedlargescale}. Resolution is marked on the top-right corner of each result in the format of \texttt{width$\times$height}. Corresponding prompts can be found in the appendix.}
    \vspace{-0.1cm}
    \label{fig:teaser}
\end{center}
}]
\begingroup
\makeatletter
\renewcommand\@makefnmark{}
\renewcommand\thefootnote{} 
\makeatother


\footnote{
$^{*}$ Equal contribution.
}
\footnote{
$^{\dagger}$ Corresponding author (\texttt{liusonghua@sjtu.edu.cn}).
}

\addtocounter{footnote}{-1}
\endgroup

\begin{abstract}

\vspace{-0.3cm}

The quadratic time and memory complexity of the attention mechanism in modern Transformer-based video generators makes end-to-end training for ultra-high-resolution videos prohibitively expensive. 
Motivated by this limitation, we introduce a training-free approach that leverages video Diffusion Transformers pre-trained at their native scale to synthesize higher-resolution videos without any additional training or adaptation. 
At the core of our method lies an inward sliding-window attention mechanism, which originates from a key observation: maintaining each query token’s training-scale receptive field is crucial for preserving visual fidelity and detail. 
However, naive local window attention, unfortunately, often leads to repetitive content and exhibits a lack of global coherence in the generated results.
To overcome this challenge, we devise a \textit{dual-path pipeline} that backs up window attention with a novel cross-attention override strategy, enabling the semantic content produced by local attention to be guided by another branch with a full receptive field and, therefore, ensuring holistic consistency. 
Furthermore, to improve efficiency, we incorporate a cross-attention caching strategy for this branch to avoid the frequent computation of full 3D attention. 
Extensive experiments demonstrate that our method delivers ultra-high-resolution videos with fine-grained visual details and high efficiency in a training-free paradigm. 
Meanwhile, it achieves superior performance on VBench, even compared to training-based alternatives, with competitive or improved efficiency. 
Codes are publicly available at \url{https://github.com/WillWu111/FreeSwim}.

\end{abstract}    
\section{Introduction}
\label{sec:intro}

Empowered by the attention mechanism’s ability to capture complex token-wise dependencies~\cite{vaswani2017attention}, Diffusion Transformers (DiTs)~\cite{peebles2023scalablediffusionmodelstransformers} have shown remarkable performance in generating high-quality images~\cite{chen2024pixartsigmaweaktostrongtrainingdiffusion, esser2024scalingrectifiedflowtransformers,gao2024luminat2xtransformingtextmodality} and videos~\cite{kong2025hunyuanvideosystematicframeworklarge, hacohen2024ltxvideorealtimevideolatent, yang2025cogvideoxtexttovideodiffusionmodels,hong2022cogvideolargescalepretrainingtexttovideo}. 
However, the quadratic complexity of attention with respect to spatial resolution makes scaling up these models computationally prohibitive, and this issue becomes even more severe in video generation, where an additional temporal dimension further amplifies the cost. 
As a result, most existing diffusion models are trained at relatively low resolutions.
This stands in stark contrast to the rapidly growing demand for ultra-high-definition content, such as 4K videos, in modern visual applications.

Considering the substantial challenges of training high-resolution video generation models, we pose a natural question in this work: \emph{Can we generate ultra-high-resolution videos using existing generators pre-trained only at lower resolution, such as 480P, without any additional training or adaptation?} 
As illustrated in Fig.~\ref{fig:motivation}(a), this is a highly non-trivial problem—directly performing inference at higher resolutions often leads to unrealistic and incoherent results.

To further investigate this issue, we take inspiration from existing training-free methods for high-resolution image generation—covering aspects such as resolution-aware attention scaling~\cite{jin2023training}, positional embedding adjustment~\cite{peng2023ntk}, denoising schedule refinement~\cite{esser2024scalingrectifiedflowtransformers}, and low-resolution guidance strategies~\cite{du2024imaxmaximizeresolutionpotential, bu2025hiflowtrainingfreehighresolutionimage}—and explore whether these techniques can be effectively extended to the video domain. 
Unfortunately, although these methods succeed in image generation, they fail to generalize effectively to video scenarios. 
As shown in Fig.~\ref{fig:motivation}(b), (c), and (d), they either remain stuck in inconsistent semantic layouts or suffer from poor fine-grained quality, counteracting the fundamental objective of ultra-high-resolution video synthesis.

\begin{figure}
    \centering
    \includegraphics[width=1\linewidth]{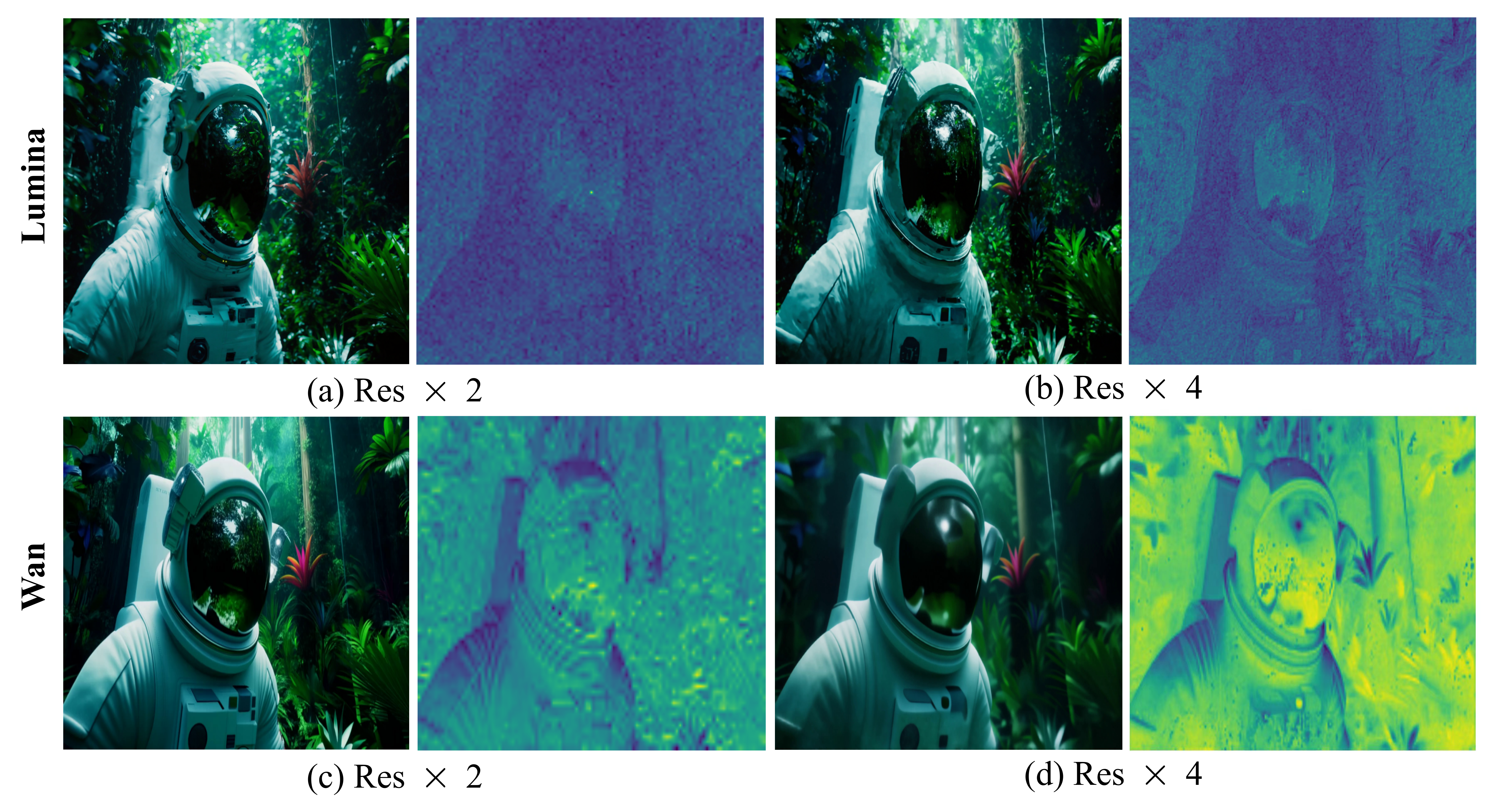}
    \vspace{-0.8cm}
    \caption{Visual results and corresponding attention maps at $\times 2$ and $\times 4$ native resolutions from Lumina-Next~\cite{gao2024luminat2xtransformingtextmodality} and Wan2.1~\cite{wan2025wanopenadvancedlargescale}, based on the same DiT architecture family.}
    \vspace{-0.5cm}
    \label{fig:attnmap}
\end{figure}

\begin{figure*}
    \centering
    \includegraphics[width=0.9\linewidth]{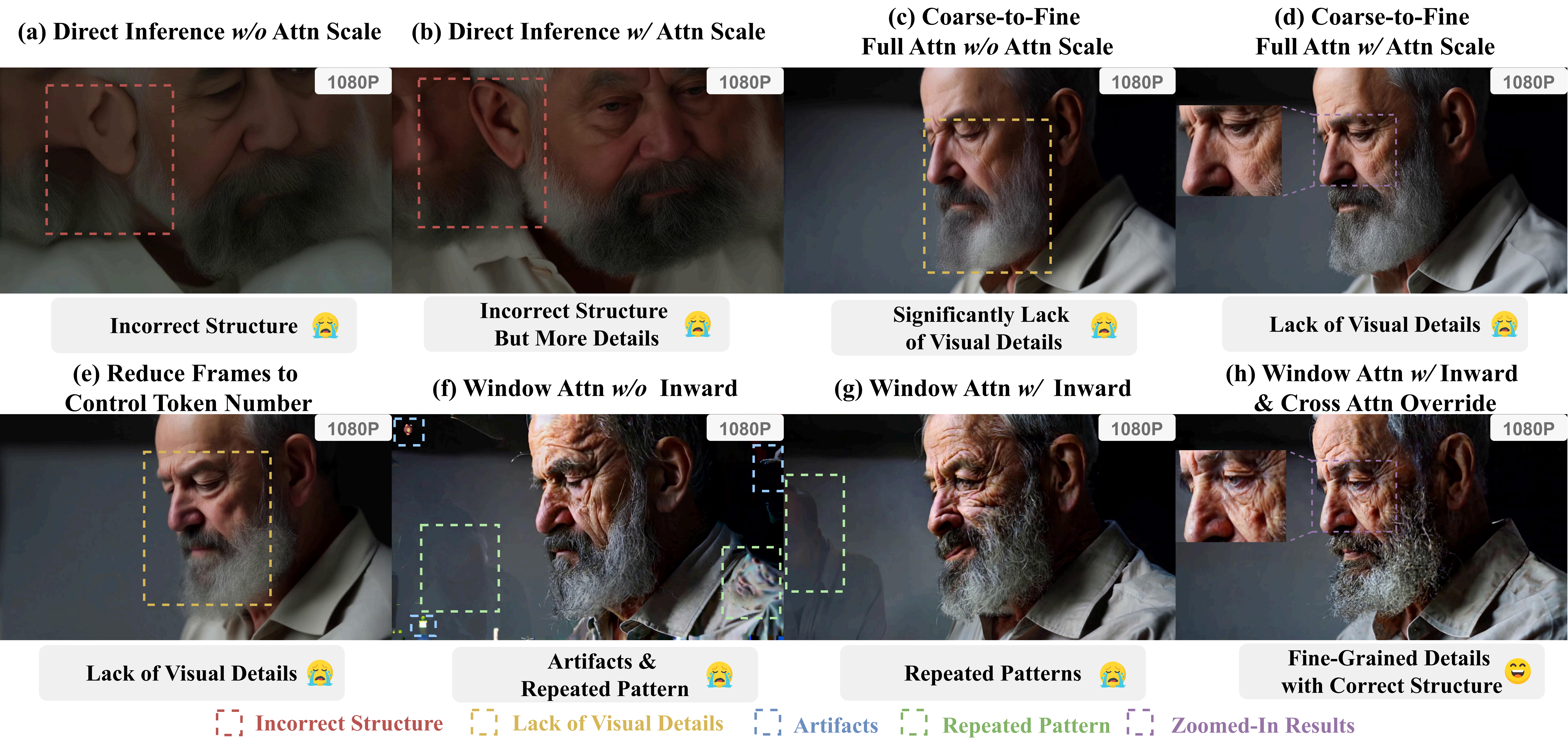}
    \vspace{-0.4cm}
    \caption{Qualitative comparison of generated results under different strategies based on Wan2.1~\cite{wan2025wanopenadvancedlargescale}. Except for (a) and (b), all other results are produced using a coarse-to-fine scheme, where a base video (832~$\times$~480) is first generated by the convention text-to-video pipeline, followed by high-resolution refinement (1920~$\times$~1088) through SDEdit~\cite{meng2021sdedit}.}
    \vspace{-0.4cm}
    \label{fig:motivation}
\end{figure*}

We therefore speculate that the unsatisfactory results stem from the limited scalability of previous methods: video generation inherently requires modeling complex 3D dependencies, making such approaches less generalizable to unseen resolution scales, even when equipped with various resolution-aware techniques. 
As illustrated in Fig.~\ref{fig:attnmap}, this effect is reflected in the attention maps: those from Lumina-Next~\cite{gao2024luminat2xtransformingtextmodality}, a DiT-based image generator, remain highly consistent across different resolution scales, whereas those from Wan2.1~\cite{wan2025wanopenadvancedlargescale}, a state-of-the-art video DiT, fail to maintain such consistency. 

Based on this analysis, the key to resolving the issue, in fact, lies in avoiding unseen generalization—specifically, preserving for each query token the same spatial receptive field that it experienced during training.
For example, in a DiT trained for 480P video generation, each query token attends only to key/value tokens within its local 480P window when synthesizing higher-resolution videos.
This forms the core of our approach: an inward sliding-window attention mechanism.
As illustrated in Fig.~\ref{fig:motivation}(g), this design substantially improves generation fidelity, producing fine-grained and visually rich details. 

Unfortunately, merely suppressing long-range interactions—similar to the dilemma faced by previous U-Net-based~\cite{ronneberger2015unetconvolutionalnetworksbiomedical} diffusion models in resolution extrapolation~\cite{qiu2025freescale,du2023demofusiondemocratisinghighresolutionimage,he2023scalecraftertuningfreehigherresolutionvisual}—inevitably results in repetitive patterns during high-resolution generation. 
Resolving this problem necessitates equipping ultra-high-resolution generation with a holistic understanding of global information.

We then turn back to the full-attention results shown in Fig.~\ref{fig:motivation}(d), which effectively preserve the overall semantic layout despite lacking fine visual details, and complement the proposed window attention approach with an additional full-attention branch. 
As shown in Fig.~\ref{fig:motivation}(h), by interacting through an override scheme at the cross-attention layers—the direct sources of semantic information—such a dual-branch pipeline effectively brings the best of both worlds, achieving both coherent semantics and visually appealing fine-grained details.

Moreover, the full-attention branch inevitably incurs substantial computational overhead.
To mitigate this, inspired by recent caching-based acceleration techniques~\cite{ma2023deepcacheacceleratingdiffusionmodels,zou2025acceleratingdiffusiontransformerstokenwise,liu2024timestep,ma2024learning}, we introduce a cross-attention feature reuse strategy that reduces the need for frequent full 3D attention computation, achieving over a $2\times$ speedup with negligible performance degradation. 

We conduct extensive experiments on modern DiT-based video generation models, \textit{e.g.}, Wan2.1~\cite{wan2025wanopenadvancedlargescale} and LTX-Video~\cite{hacohen2024ltxvideorealtimevideolatent}. 
As shown in Fig.~\ref{fig:teaser}, our method can synthesize 4K-resolution videos with strong semantic coherence and visually compelling fine-grained details. 
Quantitatively, it establishes new state-of-the-art results on VBench~\cite{huang2023vbenchcomprehensivebenchmarksuite}, even surpassing previous training-based methods in both generation quality and efficiency. 
Our contributions can be summarized as follows:
\begin{itemize}
\item We delve into the field of training-free ultra-high-resolution video generation and propose the first approach tailored for modern DiT architectures, such as Wan2.1 ~\cite{wan2025wanopenadvancedlargescale}, to the best of our knowledge. 

\item We identify the limitations of prior resolution extrapolation methods for image generation when applied to videos and propose an inward sliding-window attention mechanism complemented by a dual-path strategy to effectively overcome these issues. 

\item Extensive experiments and user evaluations confirm that our method achieves superior performance in training-free ultra-high-resolution video generation, outperforming even training-based counterparts.

\end{itemize}

\vspace{-0.05cm}
\section{Related Work}
\label{sec:Related Work}

\vspace{-0.1cm}

\subsection{Diffusion Models for Video Generation}
Diffusion models (DMs)~\cite{sohl2015deep,ho2020denoising} have been extensively explored for video generation, attracting substantial attention. Unlike conventional UNet-based video diffusion models~\cite{ho2022videodiffusionmodels,he2023latentvideodiffusionmodels,chen2023videocrafter1opendiffusionmodels,blattmann2023stablevideodiffusionscaling} that rely on convolutional backbones, Diffusion Transformer (DiT)~\cite{peebles2023scalablediffusionmodelstransformers} adopts a Transformer architecture as its core. This architectural change enables DiT to capture richer long-range dependencies and more intricate spatiotemporal relationships within video sequences. Early works~\cite{yang2025cogvideoxtexttovideodiffusionmodels, jin2025pyramidalflowmatchingefficient} demonstrated the effectiveness of spatiotemporal transformers with self-attention and a global receptive field. More recently, LTX-Video~\cite{hacohen2024ltxvideorealtimevideolatent} further optimized the interaction between the Video-VAE and the denoising transformer, while Wan2.1~\cite{wan2025wanopenadvancedlargescale} and Hunyuan~\cite{kong2025hunyuanvideosystematicframeworklarge}, both trained on large-scale video datasets, exhibited impressive performance in generating realistic videos. Although these models greatly improve generation quality, their native resolution remains limited for high-quality applications.

\vspace{-0.2cm}

\subsection{High-Resolution Visual Generation}

\vspace{-0.1cm}
In the image domain, training-free high-resolution visual generation has been extensively explored. Most existing approaches~\cite{podell2023sdxlimprovinglatentdiffusion} are built upon U-Net~\cite{ronneberger2015unetconvolutionalnetworksbiomedical}, which significantly suffers from repetitive patterns that arise from the limited local receptive fields in high-resolution synthesis.
Several works~\cite{he2023scalecraftertuningfreehigherresolutionvisual,bartal2023multidiffusionfusingdiffusionpaths,du2023demofusiondemocratisinghighresolutionimage} further expand receptive fields through dilated convolutions and fuse local and global patches to effectively suppress repetition. With the advent of recent foundational diffusion models~\cite{labs2025flux1kontextflowmatching}, DiT have become the dominant architecture, benefiting from the attention mechanism’s capacity to model complex token-wise dependencies. Training-free methods~\cite{bu2025hiflowtrainingfreehighresolutionimage,du2024imaxmaximizeresolutionpotential} on DiT are capable of maintaining global layout consistency while synthesizing fine-grained visual details.

In the video domain, high-resolution synthesis of DiT-based models introduce additional challenges, including severe computational overhead, blurring, and structural distortion. 
Some works~\cite{ye2025supergenefficientultrahighresolutionvideo} concentrate on developing the system infrastructure required for efficient high-resolution video generation. 
Most current video generation methods~\cite{ren2025turbo2k,hu2025ultragenhighresolutionvideogeneration,qiu2025cinescalefreelunchhighresolution} rely on fine-tuning with high-resolution data. In this work, we propose a fully training-free video generation framework tailored for modern DiT architectures, effectively ensuring global layout consistency while producing fine-grained details in high-resolution synthesis. A concurrent work~\cite{zhuang2025flashvsrrealtimediffusionbasedstreaming} also uses local attention for acceleration, but it does not examine its side effects, such as the severe repeated patterns that emerge at high resolutions.

\section{Method}
\begin{figure*}
    \centering
    \includegraphics[width=0.95\linewidth]{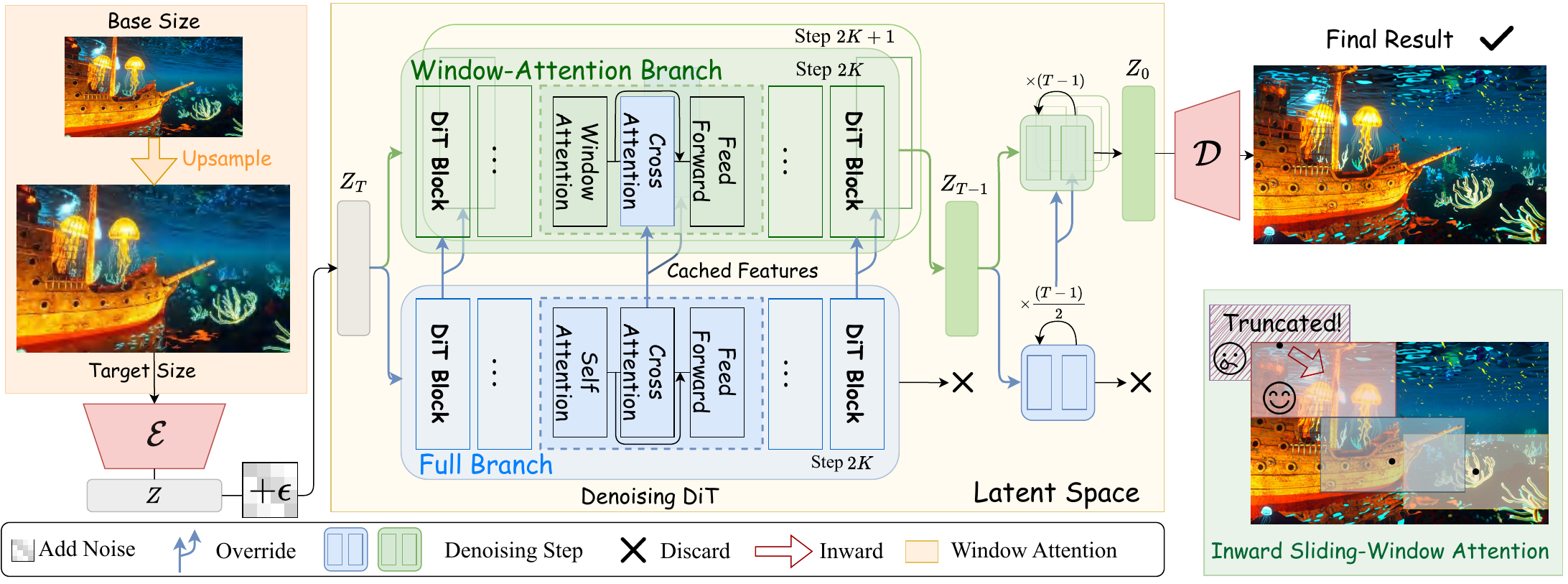}
    \vspace{-0.4cm}
    \caption{FreeSwim Framework Overview. \textbf{Upper Left:} Our parallel Dual-Path pipeline combined with cross-attention override to achieve correct global semantic structure, with the Full-Branch Feature Reuse strategy in the figure representing the case where the cross attention of Full-Branch is computed and updated every two steps ($P$=2). \textbf{Bottom Right:} Our inward window attention ensures that during inference, the spatial dimension is strictly controlled at the same scale as during training.}
    \vspace{-0.5cm}
    \label{fig:method}
\end{figure*}

\vspace{-0.2cm}

Generating 4K videos poses significant challenges, as it requires both fine-grained detail synthesis and high-level semantic coordination. 
Following prior works~\cite{du2024imaxmaximizeresolutionpotential,bu2025hiflowtrainingfreehighresolutionimage}, we adopt a coarse-to-fine generation strategy (Sec.~\ref{sec:two-stage}): first generating a video from a text prompt at the model’s native resolution and then producing a higher-resolution version based on this initial output. 
Since the first stage follows the standard inference procedure of a pre-trained text-to-video DiT, our focus in this section is on the second stage. 
The key components include an inward sliding-window attention mechanism (Sec.~\ref{sec:windowsliding}), a cross-attention override scheme (Sec.~\ref{sec:globalcrossattention}), and a feature reuse strategy for the cross-attention layers (Sec.~\ref{sec:cache}). 
An overview of our method is illustrated in Fig.~\ref{fig:method}.

\vspace{-0.05cm}

\subsection{Preliminary: Coarse-to-Fine Generation}\label{sec:two-stage}
As shown in Fig.~\ref{fig:motivation}(a), it is difficult for a text-to-video DiT pre-trained only at low resolution scales to generate high-resolution content. 
Although prior works~\cite{esser2024scalingrectifiedflowtransformers,peng2023ntk,jin2023training,du2024imaxmaximizeresolutionpotential} propose a series of insightful toolkits for resolution extrapolation, they are helpful for enhancing local details but still fail to alleviate the issue of disordered layouts in high-resolution video generation, as illustrated in Fig.~\ref{fig:motivation}(b). 
Therefore, following \cite{du2024imaxmaximizeresolutionpotential,bu2025hiflowtrainingfreehighresolutionimage}, we apply a coarse-to-fine pipeline, where the overall video layouts are handled in the low-resolution stage. 

Specifically, inspired by SDEdit~\cite{meng2021sdedit}, we first conduct text-to-video generation in the model's native resolution. 
Then, as shown in Fig.~\ref{fig:method}, this basic result is up-sampled to the target size in the pixel space and converted to the latent space through the VAE encoder. 
A random Gaussian noise is added to the encoded latents with a predefined strength (set to $0.7$ by default in our experiments), which preserves the overall content structure while perturbing fine-grained details.
The high-resolution inference process begins from this noisy state.
As shown in Fig.~\ref{fig:motivation}(c) and (d), this strategy plays a crucial role in producing reasonable content layouts and yields further performance gains when combined with the aforementioned inference toolkits.

\subsection{Inward Sliding-Window Attention}\label{sec:windowsliding}

Nevertheless, in video generation, where a model has to capture complex 3D interactions, the quality of fine-grained details remains unsatisfactory. 
In other words, previous methods effective in 2D resolution extrapolation struggle to generalize to video scenarios when the token count surpasses that observed during training. 

Based on this insight, a straightforward way to maintain the same token count $N$ as during training is to reduce the number of latent frames $F$ so that $N = H \times W \times F$, where $H$ and $W$ denote the sizes of the spatial dimensions. 
For example, using a 480P model to generate 1080P content requires reducing the frame count by $2.25\times$ to keep the token length unchanged. 
However, as shown in Fig.~\ref{fig:motivation}(e), this strategy brings limited improvement. 

We therefore hypothesize that preserving the native size of each dimension—both spatial and temporal—is essential for performance. 
Methodologically, this corresponds to a sliding-window attention mechanism: each query token attends only to key/value tokens within a local window whose receptive field matches the resolution scale used during training, \textit{e.g.}, using a 480P window for a 480P model.

Unlike classic window attention, such as Swin Transformer~\cite{liu2021swin} and Neighborhood Attention~\cite{hassani2023neighborhood}, which truncate attention windows for boundary tokens due to limited spatial context, we find it essential to maintain a consistent number of interactable key/value tokens for all queries, including those at the boundaries. 
To this end, we propose a novel inward sliding-window attention mechanism: when a query’s window extends beyond the video boundaries, we dynamically shift it inward based on the token’s spatial location, ensuring that its receptive field size matches the training resolution without truncation, as shown in Fig.~\ref{fig:method}(bottom right), which can be formulated as:
\begin{equation} \label{eq:M}
\begin{aligned}
\Delta^{(q)}_w&=\max(\frac{w}{2}-x_q,\frac{w}{2}+x_q-W+1,0),\\
\Delta^{(q)}_h&=\max(\frac{h}{2}-y_q,\frac{h}{2}+y_q-H+1,0),\\
M_{qk} &= 
\begin{cases}
1, & \text{if } \left| x_q - x_k \right| \leq \frac{w}{2} + \Delta^{(q)}_w \\
   & \text{ and } \left| y_q - y_k \right| \leq \frac{h}{2} + \Delta^{(q)}_h, \\
0, & \text{otherwise},
\end{cases}\\
O&=\mathrm{Softmax}((QK^\top)\cdot M/\sqrt{d})V,
\end{aligned}
\end{equation}
where $x_{*}$ and $y_{*}$ denote the spatial coordinates of a token; $W$ and $H$ are the target dimensions, while $w$ and $h$ are the native dimensions; $Q$, $K$, $V$, and $M$ represent the query, key, value, and attention mask matrices, respectively; $\cdot$ indicates element-wise multiplication; $d$ is the feature dimension size; and $O$ is the resulting attention output of the layer.

This inward sliding-window attention scheme is applied to all self-attention layers—the core modules for token interactions in modern DiT-based video generators like Wan2.1—in place of the original 3D full attention. 
As shown in Fig.~\ref{fig:motivation}(f) and (g), this approach significantly enhances local details, while the inward mechanism effectively eliminates boundary artifacts.

\subsection{Dual-Path with Cross-Attention Override} \label{sec:globalcrossattention}

As shown in Fig.~\ref{fig:motivation}(g), the inward sliding window attention mechanism mentioned in Sec.~\ref{sec:windowsliding}, while improving fine-grained details, inevitably produces repetitive patterns as a side effect. 
The underlying reason is the confined respective field of each token, analogous to that in previous U-Net-based generators~\cite{ronneberger2015unetconvolutionalnetworksbiomedical} like SD-1.5~\cite{rombach2022high} and SD-XL~\cite{podell2023sdxlimprovinglatentdiffusion}, where convolution serves as the primary means of feature interaction. 
A variety of methods have been developed to enable resolution extrapolation for them~\cite{qiu2025freescale,du2023demofusiondemocratisinghighresolutionimage,he2023scalecraftertuningfreehigherresolutionvisual,zhang2024hidiffusionunlockinghigherresolutioncreativity}. 

In this paper, we provide a tailored approach for DiT-based models. 
Re-examining Fig.~\ref{fig:motivation}(d) and (g), we observe a trade-off between global layout consistency and fine-grained detail quality. 
In other words, although full 3D attention produces less refined local details, it yields more coherent semantics, suggesting a potential source of global guidance in a dual-path framework that integrates both full-attention and window-attention branches. 
For the interaction between the two paths, we leverage the latent-to-text cross-attention modules, which serve as the direct source of semantic content.

Specifically, at each denoising step, we duplicate the input latent maps, concatenate them along the batch dimension, and apply full 3D attention and the proposed inward sliding-window attention to the self-attention modules, respectively. 
The cross-attention results of the latter branch, denoted as $O_{Cross}^{Window}$, are guided by the former, denoted as $O_{Cross}^{Window}$, with a strength of $\lambda$:
\begin{equation} \label{eq:cross_attention}
O_{Cross}^{Window}\leftarrow\lambda O_{Cross}^{Full}+(1-\lambda)O_{Cross}^{Window}.
\end{equation}

In practice, we find that setting $\lambda=1$ achieves the most robust performance without sacrificing fine-grained detail. 
This corresponds to an override scheme, where the cross-attention features from the window-attention branch are entirely replaced by those from the full branch. 
Please refer to Sec.~\ref{sec:ablation} for more discussion. 
As shown in Fig.~\ref{fig:motivation}(h), the proposed inward sliding-window attention, reinforced by this dual-path design, brings the best of both worlds, enabling training-free high-resolution video generation with coherent semantic structures and appealing details.


\begin{table*}[t]
\centering
\resizebox{0.97\textwidth}{!}{ 
\scriptsize  
\begin{tabular}{l | c c c c c c | c}
\toprule
\toprule
\textbf{Model} &
\makecell[c]{\textbf{Subject}\\\textbf{Consistency}} &
\makecell[c]{\textbf{Background}\\\textbf{Consistency}} &
\makecell[c]{\textbf{Motion}\\\textbf{Smoothness}} &
\makecell[c]{\textbf{Aesthetic}\\\textbf{Quality}} &
\makecell[c]{\textbf{Imaging}\\\textbf{Quality}} &
\makecell[c]{\textbf{Overall}\\\textbf{Consistency}} &
\makecell[c]{\textbf{Overall}\\\textbf{Score}} \\
\midrule
\midrule
\rowcolor{groupgray}
\multicolumn{8}{l}{\textbf{LTX-Video} ~\cite{hacohen2024ltxvideorealtimevideolatent} } \\
\midrule
CineScale (\textit{w/o} LoRA) ~\cite{qiu2025cinescalefreelunchhighresolution}       & 89.7\% & \textbf{97.3\%} & \textbf{98.7\%} & \underline{41.2\%} & 36.6\% & 16.2\% & \underline{63.3\%} \\
Ours (\textit{w/o} Cache)                                     & \underline{91.3\%} & \textbf{97.3\%} & \underline{98.5\%} & \textbf{41.3\%} & \textbf{50.0\%} & \underline{18.5\%} & \textbf{66.2\%} \\
Ours (\textit{w/} Cache, P=5)                                 & \textbf{91.4\%} & \underline{97.2\%} & \underline{98.5\%} & \textbf{41.3\%} & \underline{49.9\%} & \textbf{18.6\%} & \textbf{66.2\%} \\

\midrule
\midrule
\rowcolor{groupgray}
\multicolumn{8}{l}{\textbf{Wan2.1 14B} ~\cite{wan2025wanopenadvancedlargescale}} \\
\midrule
CineScale (\textit{w/o} LoRA) ~\cite{qiu2025cinescalefreelunchhighresolution}        & \textbf{96.3\%} & \textbf{96.5\%} & \textbf{98.8\% }& \underline{59.9\%} & \underline{65.1\%} & \underline{23.8\%} & \underline{73.4\%} \\
Ours (\textit{w/o} Cache)                                     & \underline{95.7\%} & \underline{96.4\%} & \underline{98.3\%} & \textbf{60.9\% }& \textbf{65.8\%} & \textbf{24.6\%} & \textbf{73.6\%} \\
\midrule
\midrule
\rowcolor{groupgray}
\multicolumn{8}{l}{\textbf{Wan2.1 1.3B} ~\cite{wan2025wanopenadvancedlargescale}} \\
\midrule
Real-ESRGAN ~\cite{wang2021realesrgantrainingrealworldblind}  & 94.7\%  & 96.4\%  & 97.9\%  & 55.9\%  & \underline{63.9\%}  & 24.8\%  & 72.3\%  \\
Upscale-A-Video ~\cite{zhou2023upscaleavideotemporalconsistentdiffusionmodel}
                                                             & 91.0\% & 97.0\% & 97.2\% & 55.6\% & 62.9\% & 22.9\% & 71.1\% \\
\midrule
I-Max ~\cite{du2024imaxmaximizeresolutionpotential}           & 92.8\% & 96.9\% & 95.3\% & \underline{59.6\%} & 61.1\% & 23.0\% & 71.5\% \\
HiFlow ~\cite{bu2025hiflowtrainingfreehighresolutionimage}    & 90.7\% & 96.3\% & 95.2\% & 54.8\% & 54.4\% & 24.7\% & 69.4\% \\
CineScale (\textit{w/o} LoRA) ~\cite{qiu2025cinescalefreelunchhighresolution}        & 94.3\% & 96.9\% & \textbf{98.2\%} & 45.9\% & 40.9\% & 16.8\% & 65.5\% \\
CineScale (\textit{w/} LoRA) ~\cite{qiu2025cinescalefreelunchhighresolution}         & 93.9\% & 96.8\% & 97.4\% & 56.7\% & 60.7\% & 23.5\% & 71.5\% \\
\midrule
Ours (\textit{w/o} Cache)                                     & 94.7\% & 97.2\% & 97.7\% & 57.9\% & 63.4\% & \underline{25.1\%} & 72.7\% \\
Ours (\textit{w/o} CFG-Full)                                       & \textbf{97.1\%} & \textbf{98.2\%} & \textbf{98.2\%} & 58.6\% & 61.8\% & \underline{25.1\%} & \underline{73.2\%} \\
Ours (\textit{w/} Cache, P=2)                                 & 95.1\% & \underline{97.6\%} & 97.9\% & \textbf{61.2\%} & \textbf{65.3\%} & 24.9\% & \textbf{73.7\%} \\
Ours (\textit{w/} Cache, P=5)                                 & 94.4\% & 97.4\% & 97.8\% & 56.2\% & 62.6\% & \textbf{25.2\%} & 72.3\% \\
Ours (\textit{w/} Cache, P=8)                                 & \underline{95.2\%} & \underline{97.6\%} & \underline{98.0\%} & 54.3\% & 61.0\% & 22.0\% & 71.4\% \\
\bottomrule
\bottomrule
\end{tabular}
}
\vspace{-0.3cm}
\caption{Video comparison with both training-based and training-free models at 1080P. Our method FreeSwim achieves the best scores across almost all metrics and ranks first overall in the Wan Series~\cite{wan2025wanopenadvancedlargescale} and LTX Series~\cite{hacohen2024ltxvideorealtimevideolatent} respectively. The highest value is \textbf{bold}, and the second-highest is \underline{underlined}.}
\label{tab:1080p}
\vspace{-0.5cm}

\end{table*}

\vspace{-0.25cm}

\subsection{Full-Branch Feature Cache and Reuse}\label{sec:cache}

By default, at each denoising step, a latent map must pass through the full-attention branch to compute cross-attention guidance for the window-attention branch, which introduces substantial computational overhead, especially at ultra-high-resolution scales like 4K. 
To address this, motivated by prior work on caching for diffusion models~\cite{ma2023deepcacheacceleratingdiffusionmodels,ma2024learning,liu2024timestep,zou2025acceleratingdiffusiontransformerstokenwise}, we propose a step-wise reuse strategy for the cross-attention features from the full branch. 

Specifically, we update and cache the cross-attention guidance from the full-attention branch every $P$ steps. 
The cached results are then reused to override the cross-attention outputs of the window-attention branch in the intervening steps, which has minimal impact on video quality when $P$ is set to $2\sim8$ according to our experiments. 
Fig.~\ref{fig:method} presents the case of $P=2$. 

Moreover, for models using classifier-free guidance~\cite{ho2022classifierfreediffusionguidance}, we find it unnecessary to apply the dual-path pipeline to the unconditional branch. 
In other words, we omit full-attention guidance for this branch and rely solely on inward sliding-window attention in the self-attention layers. 
Together, these two designs achieve over a more than $2\times$ inference speedup at 1080P resolution.

\definecolor{headergray}{HTML}{F2F2F2}
\definecolor{groupgray}{HTML}{FAFAFA}
\definecolor{ourscache}{HTML}{E3F2FD}
\definecolor{oursnocache}{HTML}{FFF3E0}

\vspace{-0.05cm}

\section{Experiments}
\subsection{Settings and Implementation Details}

In this part, we conduct experiments using Wan2.1~\cite{wan2025wanopenadvancedlargescale} and LTX-Video~\cite{hacohen2024ltxvideorealtimevideolatent}, known for their state-of-the-art performance in video generation. Wan2.1~\cite{wan2025wanopenadvancedlargescale} includes both a 1.3B version trained solely at $832 \times 480$ (480P) resolution and a 14B version trained on a mixed dataset of 480P and $1280 \times 720$ (720P). 
LTX-Video~\cite{hacohen2024ltxvideorealtimevideolatent} is trained solely on $512 \times 768$. 
More results from other DiT models are presented in the Appendix. 
We utilize FlexAttention~\cite{dong2024flexattentionprogrammingmodel} in PyTorch~\cite{paszke2019pytorchimperativestylehighperformance}, which provides efficient low-level optimization for sparse attention, to implement the proposed inward sliding-window attention.

We evaluate the performance of our method using VBench~\cite{huang2023vbenchcomprehensivebenchmarksuite}, a benchmark designed to assess both visual quality and semantic coherence. For $1920 \times 1088$ (1080P) video generation, we randomly select 60 prompts from the standard prompt suite of VBench~\cite{huang2023vbenchcomprehensivebenchmarksuite} and follow its official evaluation settings. Each method generates five videos per prompt using five different random seeds. We then employ the official VBench metrics to ensure a fair comparison among all methods. For $3380 \times  1920$ (3K), we adopt 20 randomly selected prompts for evaluation.

\begin{figure*}[ht]
    \centering
    \includegraphics[width=1\linewidth]{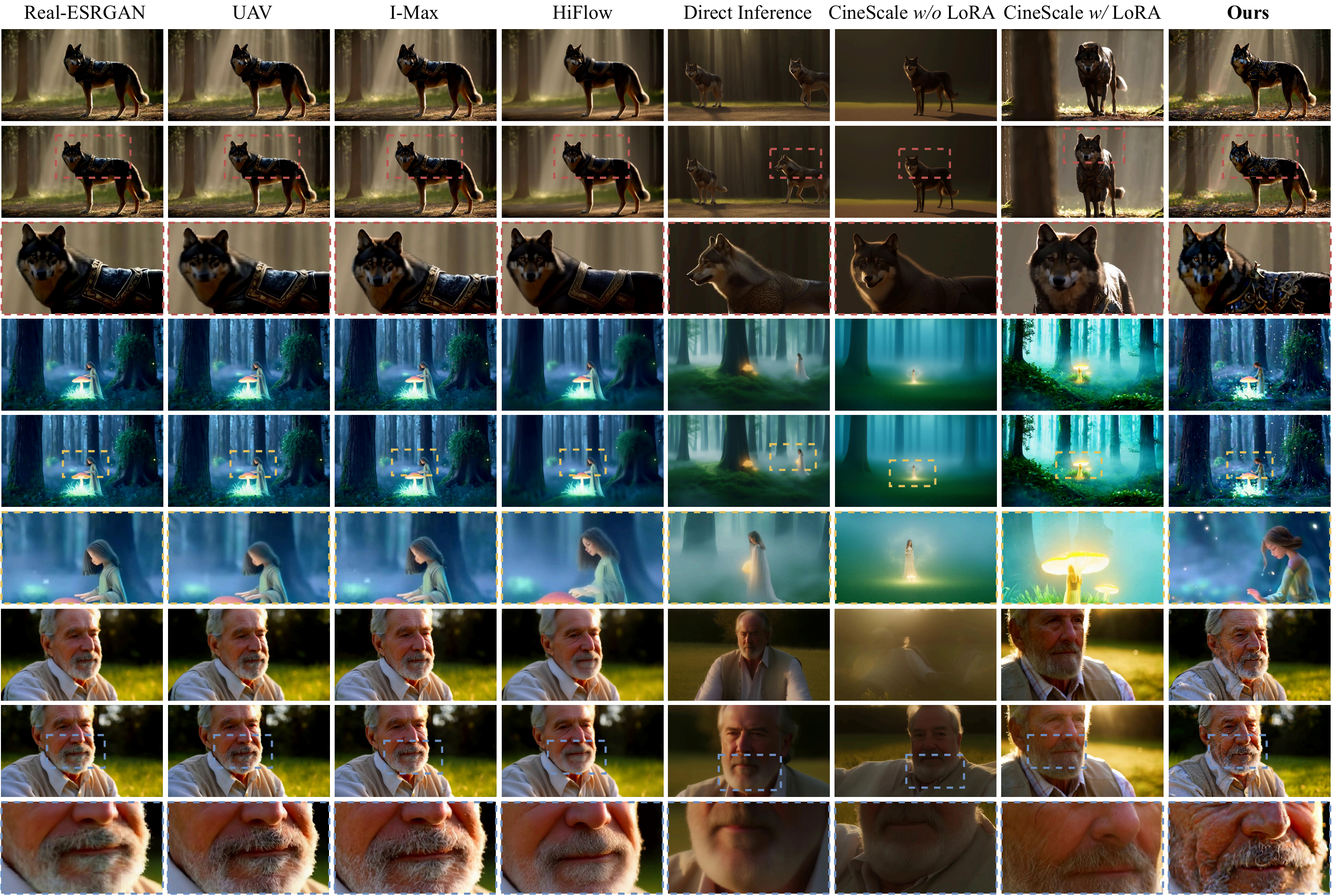}
    \caption{Visual comparison of synthesized 1080P videos for models based on Wan. Our method FreeSwim yields high-resolution videos characterized by high-fidelity details and coherent structure. Best viewed zoomed in. Corresponding prompts can be found in the appendix.}
    \label{fig:case study}
\end{figure*}


\subsection{Main Comparisons}

We evaluate our method on LTX-Video and Wan2.1. 
We consider the following baselines: (1) \textbf{Low-Level Super Resolution Methods}, including Real-ESRGAN~\cite{wang2021realesrgantrainingrealworldblind} and Upscale-A-Video~\cite{zhou2023upscaleavideotemporalconsistentdiffusionmodel}, whose base videos are derived from Wan2.1~\cite{wan2025wanopenadvancedlargescale}; (2) \textbf{Training-Free High-Resolution Generation Methods}, including I-Max~\cite{du2024imaxmaximizeresolutionpotential}, HiFlow~\cite{bu2025hiflowtrainingfreehighresolutionimage}, and CineScale~\cite{qiu2025cinescalefreelunchhighresolution} without LoRA, \textit{i.e.}, a coarse-to-fine strategy based on \cite{meng2021sdedit} that is also adopted in our approach; and (3) \textbf{Training-Based High-Resolution Generation Methods}, including CineScale~\cite{qiu2025cinescalefreelunchhighresolution} with LoRA~\cite{hu2021loralowrankadaptationlarge}, which conducts adaptation at 2K resolution.

\begin{table}[t]
\centering
\setlength{\tabcolsep}{1.1pt}
\resizebox{\linewidth}{!}{
\renewcommand{\arraystretch}{1.18}
\begin{tabular}{l|cccc|c|c}
\toprule
\textbf{Model} &
\textbf{V1} &
\textbf{V2} &
\textbf{V3} &
\textbf{V4} &
\textbf{V5} &
\textbf{V6} \\
\midrule

{CineScale (\textit{w/o} LoRA)}~\cite{qiu2025cinescalefreelunchhighresolution} & \textbf{98.8\%} & 42.1\% & 29.9\% & 13.5\% & 46.1\% & 119 min 43 s \\

{CineScale (\textit{w/} LoRA)}~\cite{qiu2025cinescalefreelunchhighresolution} & \underline{96.7\%} & 44.7\% & 52.2\% & 17.5\% & 52.8\% & \underline{119 min 41 s} \\

{Ours (\textit{w/o} Cache)}  & \underline{96.7\%} & \textbf{56.4\%} & \textbf{59.3\%} & \textbf{22.5\%} & \textbf{58.7\%} & 288 min 10 s \\

{Ours (\textit{w/} Cache, P=5)}  & 96.0\% & \underline{49.1\%} & \underline{57.3\%} & \underline{21.4\%} & \underline{56.0\%} & \textbf{114 min 6 s} \\

\bottomrule
\end{tabular}
}
\vspace{-0.3cm}
\caption{Video comparison with both training-based and training-free variants of CineScale~\cite{qiu2025cinescalefreelunchhighresolution} at 3K. V1: Background Consistency. V2: Aesthetic Quality. V3: Imaging Quality. V4: Overall Consistency. V5: Overall Scores. V6: Inference Time.}
\label{tab:3k}
\vspace{-0.7cm}
\end{table}

As shown in Tabs.~\ref{tab:1080p} and~\ref{tab:3k}, our method demonstrates superior performance across both 1080P and 3K resolutions, particularly in terms of aesthetic quality (including layout, color richness, and harmony), imaging quality (capturing distortions such as over-exposure, noise, and blur), and overall consistency (reflecting both semantic and style alignment). These results validate the effectiveness of our method, enhancing fine-grained aesthetic details and improving the global-layout accuracy of the generated videos. Our method remains competitive to the baselines on other metrics and ranks first on the overall scores.

To visually demonstrate the superiority of our approach, as illustrated in Fig.~\ref{fig:case study}, we compare our method with the baselines ~\cite{bu2025hiflowtrainingfreehighresolutionimage,du2024imaxmaximizeresolutionpotential,zhou2023upscaleavideotemporalconsistentdiffusionmodel,wang2021realesrgantrainingrealworldblind,qiu2025cinescalefreelunchhighresolution} designed for high-resolution visual generation on Wan2.1-1.3B~\cite{wan2025wanopenadvancedlargescale}. Under the same prompt, our method consistently produces the most visually appealing results, featuring the richest fine-grained details and the highest semantic consistency with the prompt. In particular, in the second example, our generated video vividly presents the prompt-specific details, such as the facial details, which are barely manifested in the results of other approaches.

\definecolor{headergray}{RGB}{245,245,245}   
\definecolor{groupgray}{RGB}{250,250,250}    
\definecolor{baseline}{RGB}{248,248,248}     
\definecolor{ourscache}{HTML}{E8F5E9}       
\definecolor{oursnocache}{HTML}{E3F2FD}      
\definecolor{darkgreen}{RGB}{0,100,0}

\begin{table*}[t]
\centering
\setlength{\tabcolsep}{1.1pt}
\resizebox{\textwidth}{!}{
\renewcommand{\arraystretch}{1.18}
\begin{tabular}{l|c|c|c|c|cccccc|c|cc}
\toprule

\textbf{Model} &
\makecell[c]{\textbf{Window}\\\textbf{Self Attn}} &
\makecell[c]{\textbf{Cross Attn}\\\textbf{Override}} &
\makecell[c]{\textbf{\textit{w/o} CFG}\\\textbf{-Full}} &
\makecell[c]{\textbf{Cache \&}\\\textbf{Reuse}} &
\makecell[c]{\textbf{Subject}\\\textbf{Consistency}} &
\makecell[c]{\textbf{Background}\\\textbf{Consistency}} &
\makecell[c]{\textbf{Motion}\\\textbf{Smoothness}} &
\makecell[c]{\textbf{Aesthetic}\\\textbf{Quality}} &
\makecell[c]{\textbf{Imaging}\\\textbf{Quality}} &
\makecell[c]{\textbf{Overall}\\\textbf{Consistency}} &
\makecell[c]{\textbf{Overall}\\\textbf{Score}} &
\makecell[c]{\textbf{Inference}\\\textbf{Time}} \\
\midrule

\textbf{Wan-Only Full}  &  \xmark &  \xmark &\xmark  & \xmark & 94.3\% & 96.9\% & 98.2\% & 45.9\% & 40.9\% & 16.8\% & 65.5\% & 66 min 46 s \\

\textbf{Wan-Only Window} &\cmark & \xmark  & \xmark & \xmark & 94.6\% & 97.2\% & 97.5\%  & 58.2\% & 61.9\% & 25.3\% & 72.5\% \small{\textcolor{darkgreen}{+7.0\%}}  & 32 min 29 s \\

\textbf{Ours (\textit{w/o} Cache)} & \cmark & \cmark & \xmark &  \xmark & 94.7\% & 97.2\% & 97.7\%  & 57.9\% & 63.4\% & 25.1\% & 72.7\% \small{\textcolor{darkgreen}{+7.2\%}} & 79 min 15 s \\

\textbf{Ours (\textit{w/o} CFG-Full)} & \cmark & \cmark & \cmark &  \xmark & 97.1\% & 98.2\% & 98.2\% &  58.6\% &  61.8\% & 25.1\% &  73.2\% \small{\textcolor{darkgreen}{+7.7\%}} & 64 min 5 s \small{\textcolor{darkgreen}{$1.2\times$}} \\

\textbf{Ours (\textit{w/} Cache, P=2)}& \cmark & \cmark & \cmark & \cmark &  95.1\% & 97.6\% & 97.9\% & 61.2\% & 65.3\% & 24.9\% &  73.7\% \small{\textcolor{darkgreen}{+8.2\%}} & 54 min 40 s \small{\textcolor{darkgreen}{$1.5\times$}} \\

\bottomrule
\end{tabular}
}
\vspace{-0.3cm}
\caption{Ablation study on the Wan2.1-1.3B~\cite{wan2025wanopenadvancedlargescale} at 1080P resolution. Higher is better for all metrics except inference time.}
\label{tab:ablation_wan}
\vspace{-0.3cm}

\end{table*}

\begin{figure*}[ht]
    \centering
    
    \begin{subfigure}{0.73\linewidth}
        \centering
        \includegraphics[width=\linewidth,height=3.5cm]{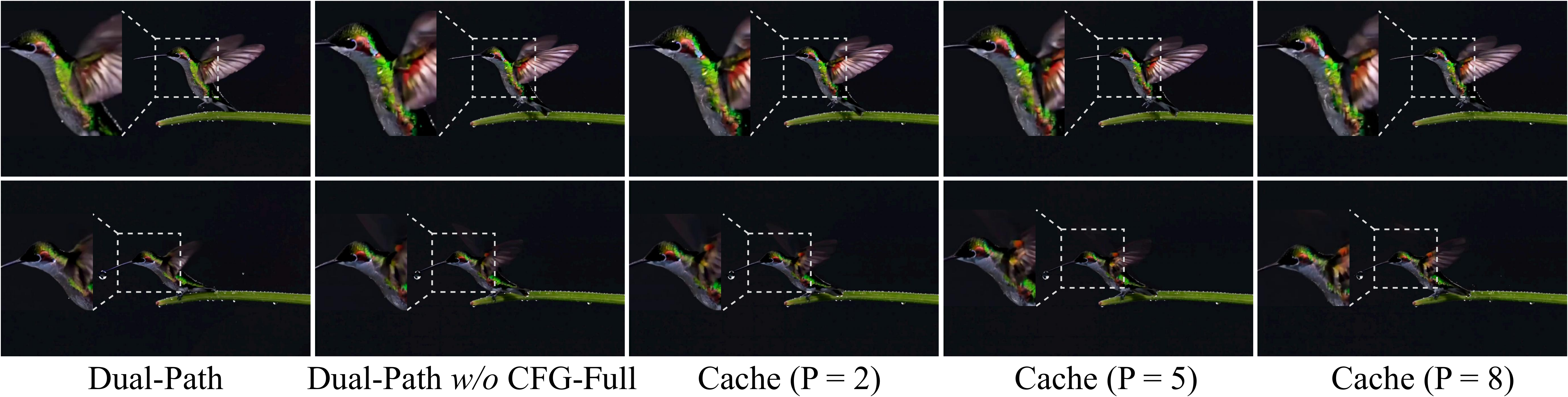}
        \label{fig:cache}
    \end{subfigure}
    \hfill
    \begin{subfigure}{0.26\linewidth}
        \centering
        \includegraphics[width=\linewidth,height=3.5cm]{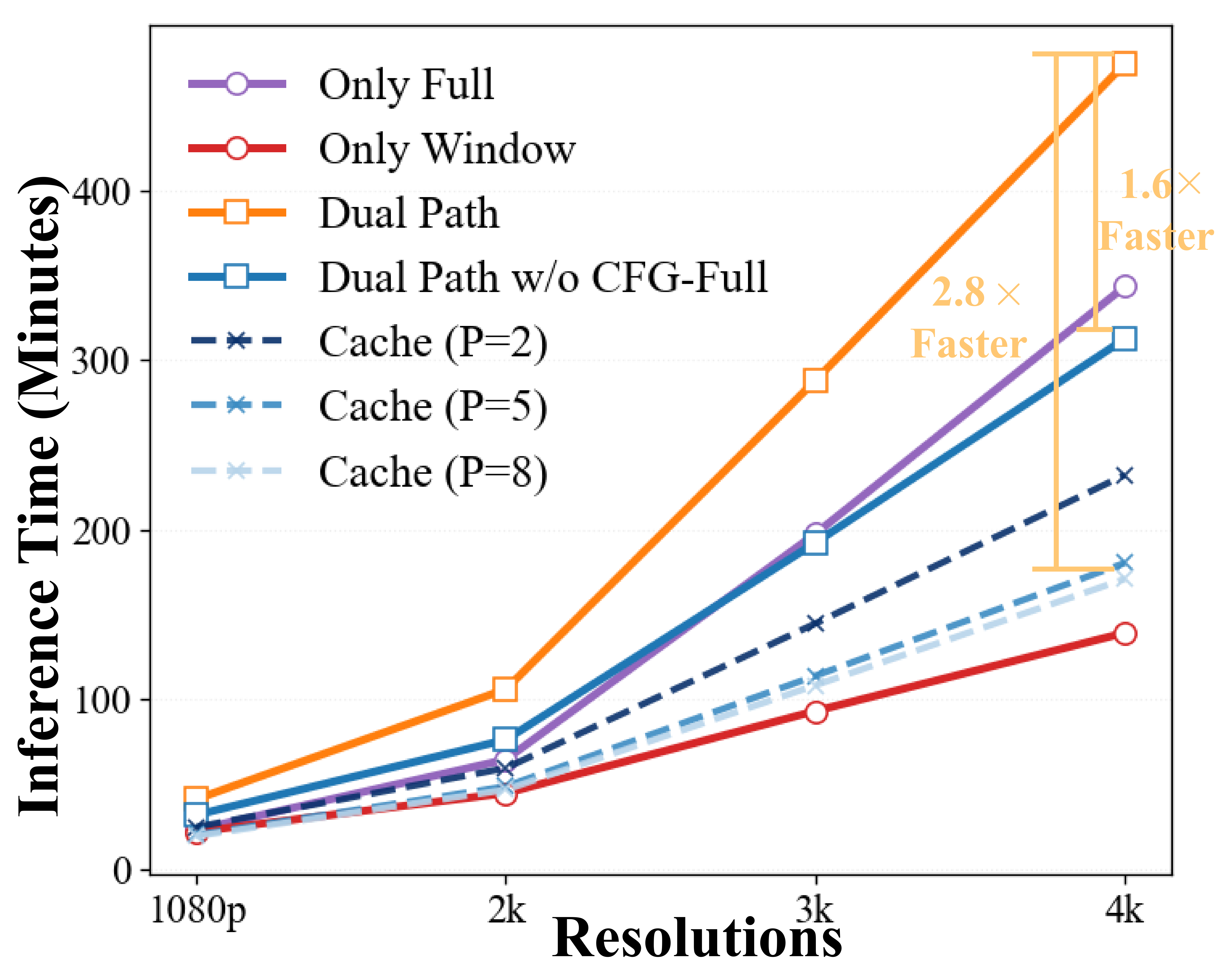}
        \label{fig:time}
    \end{subfigure}
    \vspace{-0.7cm}
    \caption{\textbf{Left:} Five processing pipelines are presented, where the inference speed increases from left to right. However, visual details are preserved without loss, ensuring a consistent global layout throughout. \textbf{Right:} Impact of different processing pipelines on inference time.}
    \vspace{-0.3cm}
    \label{fig:models}
\end{figure*}

\vspace{-0.1cm}

\subsection{User Study}
\vspace{-0.1cm}
To further qualitatively evaluate the performance of our method, we conduct a human study to assess the subjective aesthetic perception of the generated videos. Specifically, we compare four methods, all based on Wan2.1-1.3B~\cite{wan2025wanopenadvancedlargescale}: the training-based approach CineScale ~\cite{qiu2025cinescalefreelunchhighresolution} (w/ LoRA), the training-free variant CineScale ~\cite{qiu2025cinescalefreelunchhighresolution} (w/o LoRA), the direct inference results of the original model, and our proposed method. All methods are evaluated using the same prompts from VBench ~\cite{huang2023vbenchcomprehensivebenchmarksuite}. During the study, each participant was presented with the generated videos in a randomized order and expected to select the best video based on three criteria: aesthetic appeal, detail richness, and text alignment.

\begin{figure}[!t]
    \centering
    \includegraphics[width=1\linewidth]{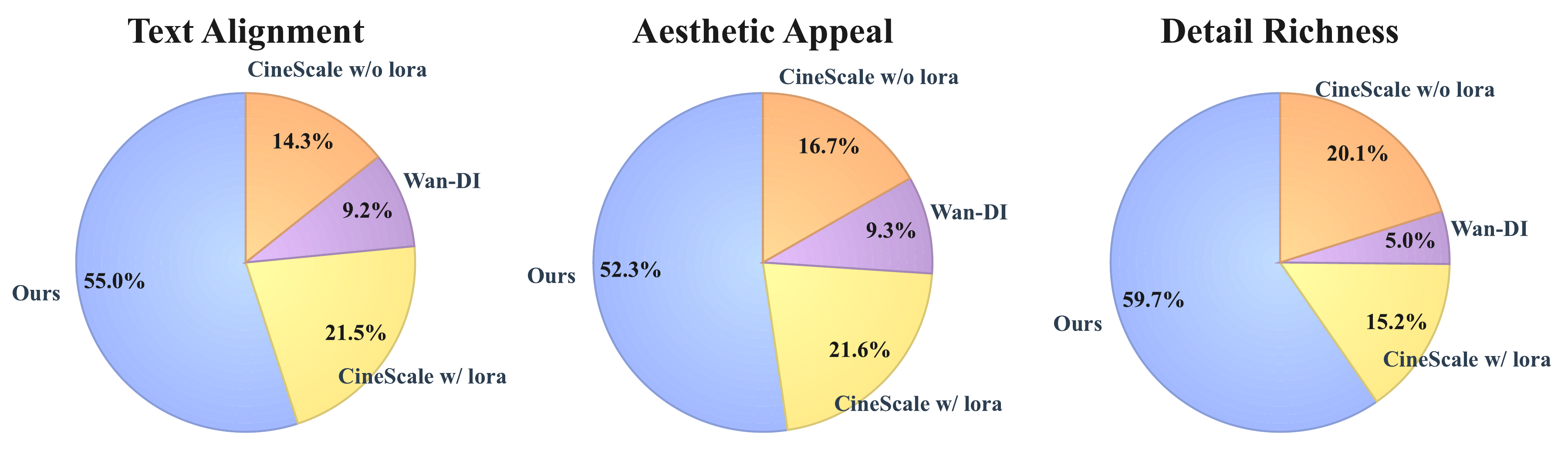}
    \vspace{-0.7cm}
    \caption{Results of user study for high-resolution video generation. Participants were expected to select the best method based on text alignment, coverage quality, and video quality.}
    \label{fig:user study}

    \vspace{-0.4cm}
\end{figure}

In total, 29 participants took part in the study and provided their subjective preferences. As shown in Fig.~\ref{fig:user study}, our method consistently receives the highest preference across all the metrics.

\vspace{-0.2cm}

\subsection{Ablation Study}\label{sec:ablation}
\vspace{-0.1cm}
\textbf{Contribution of Proposed Strategy.} To validate the effectiveness of our proposed method, we conduct detailed ablation studies on Wan2.1-1.3B~\cite{wan2025wanopenadvancedlargescale} as shown in Tab. \ref{tab:ablation_wan}. First, we demonstrate that incorporating \textit{Inward Sliding-Window Attention} significantly enhances visual details compared to direct inference, resulting in substantial improvements in multiple metrics, including aesthetic quality and image quality. To address the repetition problem caused by window attention, the proposed \textit{Dual-Path with Cross-Attention override} strategy further improves semantic consistency in the generated videos, enhancing both imaging quality and overall score. Additionally, to mitigate the computational overhead of the dual-path design, we adopt a \textit{Full-Branch Feature Cache and Reuse} strategy, which not only significantly accelerates inference but also delivers a $1.0\%$ improvement over the variants of \textit{w/o} cache.

\textbf{Effect of the \texorpdfstring{$\lambda$}{lambda} Coefficient.} As a control factor of cross-attention strength between branches, $\lambda$ in Eq.~\ref{eq:cross_attention} plays a crucial role in balancing global layout consistency and fine-grained detail quality. To better understand its influence, we conduct a detailed analysis of the hyperparameter $\lambda$ in the override strategy.

\begin{figure}[!t]
    \centering
    \includegraphics[width=1.0\linewidth]{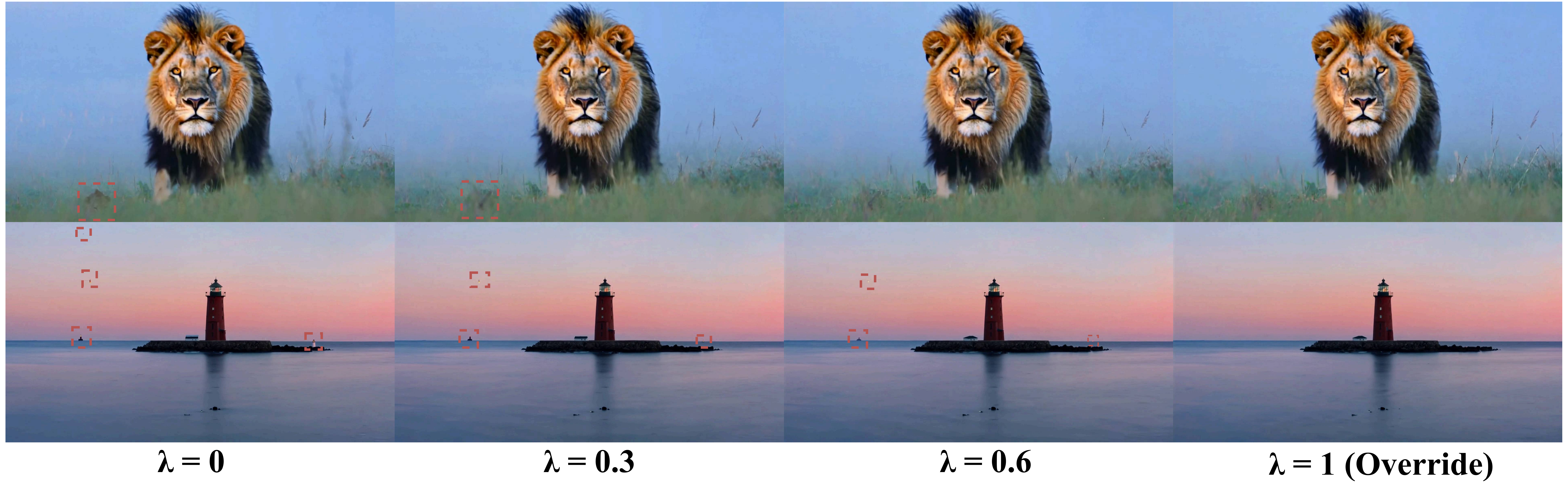}
    \caption{The effects of $\lambda$ under different settings. Repeated patterns are highlighted with red boxes.}
    \label{fig:lambda}
    \vspace{-0.3cm}
\end{figure}

As shown in Fig.~\ref{fig:lambda}, on the one hand, increasing $\lambda$ mitigates repetition, but no single value works across different cases. At $\lambda=0.3$, both examples remain repetitive; at $\lambda=0.6$, the first exhibits no repeated lions, while the second still shows duplicated beacons. 
On the other hand, the quality of fine-grained details remains nearly unchanged across different values of $\lambda$. 
We therefore set $\lambda=1$, which exhibits the best robustness across all scenarios and corresponds to our \textit{cross-attention override} strategy.

\textbf{Exploration of Reusing Mechanisms.} 
We study the effect of the proposed Full-Branch Feature Reuse strategy in this part. 
As shown in Fig.~\ref{fig:models}, we first observe that the unconditional branch does not benefit from the dual-path strategy. Removing it reduces computation and yields a $1.6\times$ speedup over the dual-path baseline, and even surpasses direct inference in efficiency. We then study the recompute interval $P$ of cross-attention from the full branch. Increasing $P$ lowers the refresh frequency and increases feature reuse, thereby improving inference efficiency. At higher resolutions, the computational cost of the full branch grows, so this caching and reusing strategy plays a more crucial role in efficiency. At 4K resolution, $P=8$ achieves up to a $2.8\times$ speedup. To balance visual quality and efficiency, we adopt $P=2$ by default, yielding a $1.5\times$ speedup. 
The effectiveness of this reuse and caching strategy is further confirmed in the quantitative results in Tabs.~\ref{tab:1080p}(bottom) and \ref{tab:ablation_wan}.

\vspace{-0.1cm}
\section{Conclusion}
In this paper, we introduce FreeSwim, a novel training-free paradigm aimed at ultra-high-resolution video generation by leveraging pre-trained text-to-video diffusion transformers trained only at low resolution scales. By introducing an inward sliding-window attention scheme and a window-attention and full-attention dual-path approach with cross-attention override, FreeSwim effectively addresses the underlying challenges in high-resolution generation, such as repetitive patterns and quality degradation, without the need for any data collection or model training. Moreover, our method supports accelerating inference by caching and reusing cross-attention from the full-attention branch, resulting in negligible performance degradation. Experimental results demonstrate the superiority and efficiency of FreeSwim in ultra-high-resolution video generation, surpassing existing methods in video quality. 

{
    \small
    \bibliographystyle{ieeenat_fullname}
    \bibliography{main}
}

\clearpage
\setcounter{page}{1}
\maketitlesupplementary



\begin{figure*}[ht]
    \centering
    \vspace{-0.8cm}
    \includegraphics[width=\textwidth]{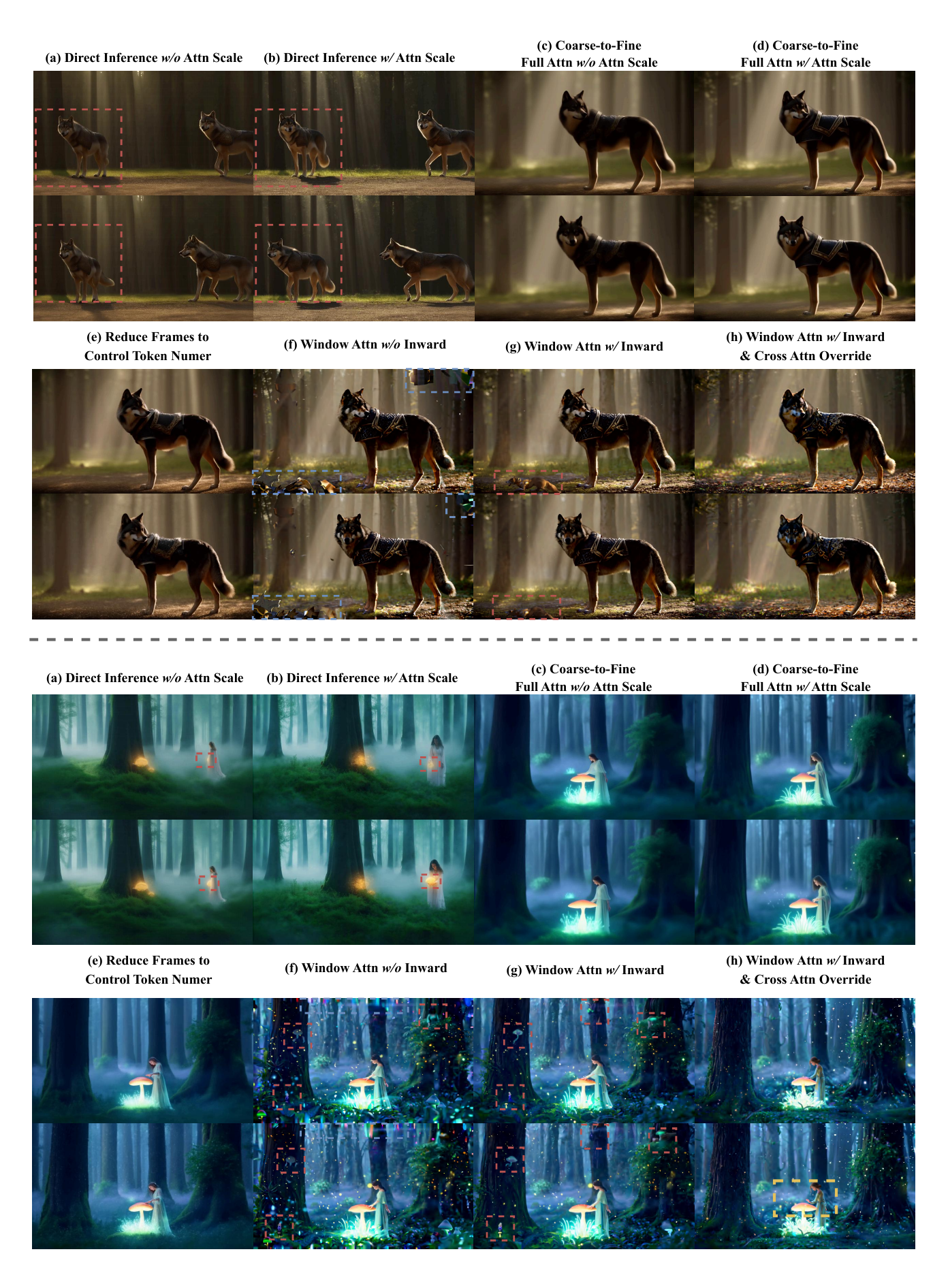}
    \vspace{-1cm}
    \caption{This figure presents a comprehensive ablation study in Fig.~\ref{fig:case study}, with red boxes highlighting repetitive content and blue boxes highlighting artifacts.}
    \label{fig:more ablation}
\end{figure*}

\textbf{Ablation Study:} 
As shown in Fig.~\ref{fig:more ablation}, we extend the motivation example in Fig.~\ref{fig:motivation} with more cases to conduct a comprehensive ablation study, further validating the effectiveness of our method. By comparing Fig.~\ref{fig:more ablation}(a)(b) and (c)(d), we observe that the proposed coarse-to-fine design preserves a coherent global semantic structure. Then, the comparison between Fig.~\ref{fig:more ablation}(e) and (f) shows that roughly controlling the receptive field, as same as the training scale, can enhance local visual details but leads to boundary artifacts and severe repetition. By comparing Fig.~\ref{fig:more ablation}(f) and (g), we further observe that enforcing a uniform interaction range for each query is crucial: it alleviates the insufficient receptive field of boundary queries and effectively suppresses phantom artifacts, although repetitive patterns still remain. Finally, the comparison between Fig.~\ref{fig:more ablation}(g) and (h) demonstrates that the full-branch cross-attention override significantly mitigates repetitive patterns caused by the local receptive field of window attention, providing strong empirical evidence for the overall effectiveness of our approach.

\begin{figure*}
    \centering
    \includegraphics[width=\textwidth]{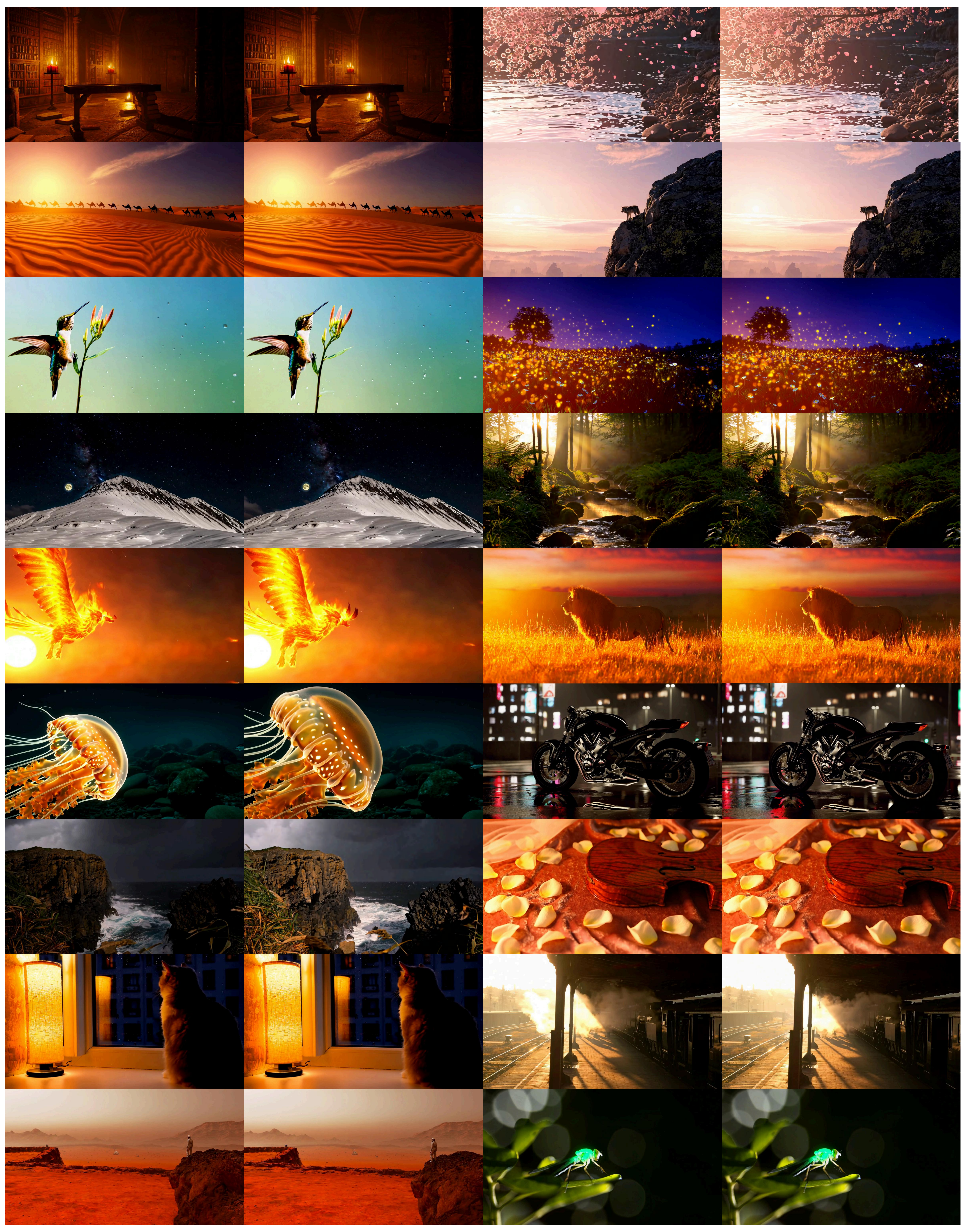}
    \caption{Qualitative results of 2K by our method FreeSwim.}
    \label{fig:case supp}
\end{figure*}

\textbf{Case Study:} 
As shown in Fig.~\ref{fig:case supp}, we adapt FreeSwim to Wan2.1~\cite{wan2025wanopenadvancedlargescale} to generate $2$K videos. These qualitative results demonstrate that our method can consistently preserve coherent global semantic structure while producing fine-grained details across diverse scenarios, including landscapes, objects, and both static and dynamic scenes.

\begin{figure*}
    \centering
    \includegraphics[width=\linewidth]{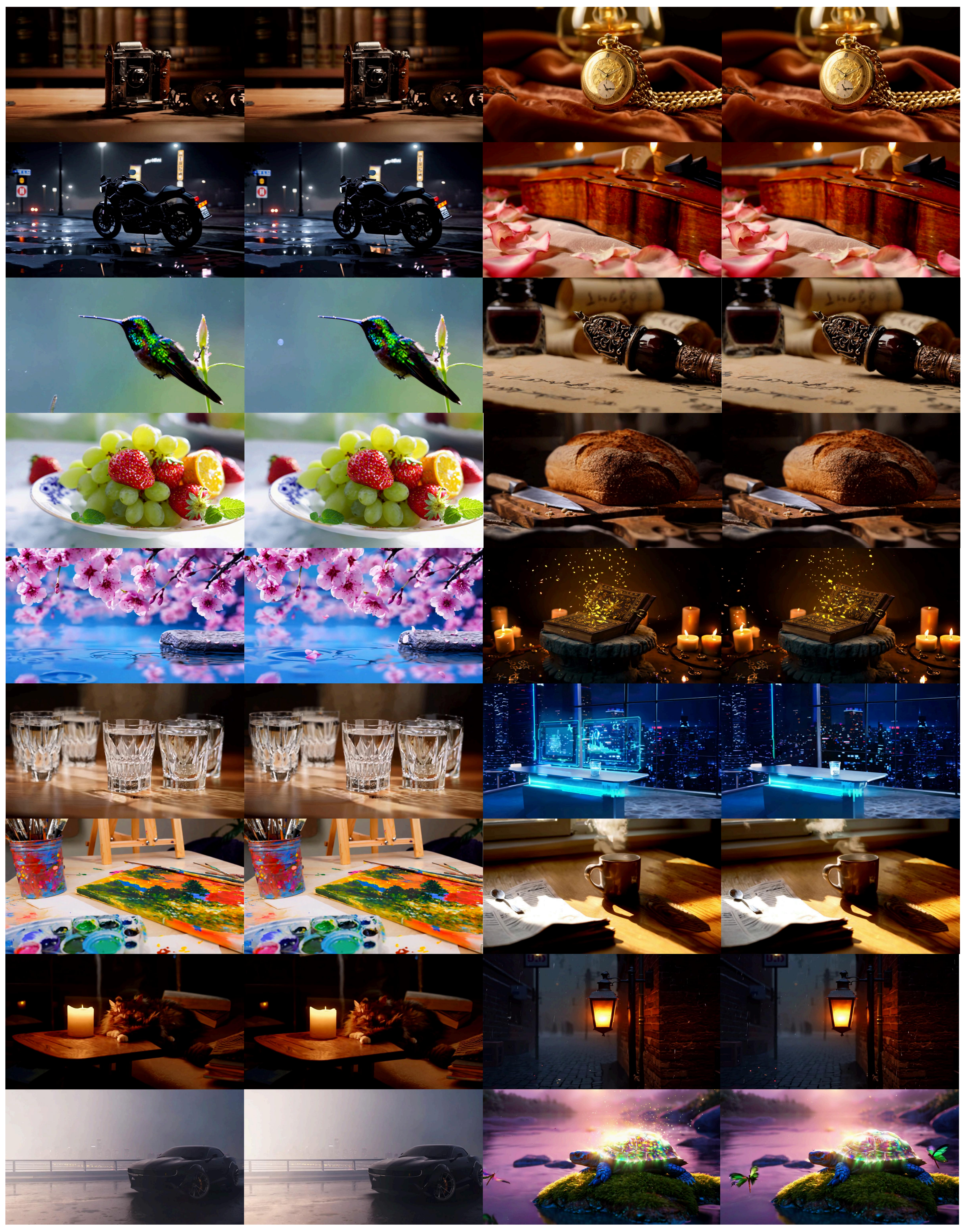}
    \caption{Qualitative results of 1080P generated by our method FreeSwim integrated into \textbf{VACE} pipelines.}
    \label{fig:vace}
\end{figure*}

\begin{figure*}
    \centering
    \includegraphics[width=\linewidth]{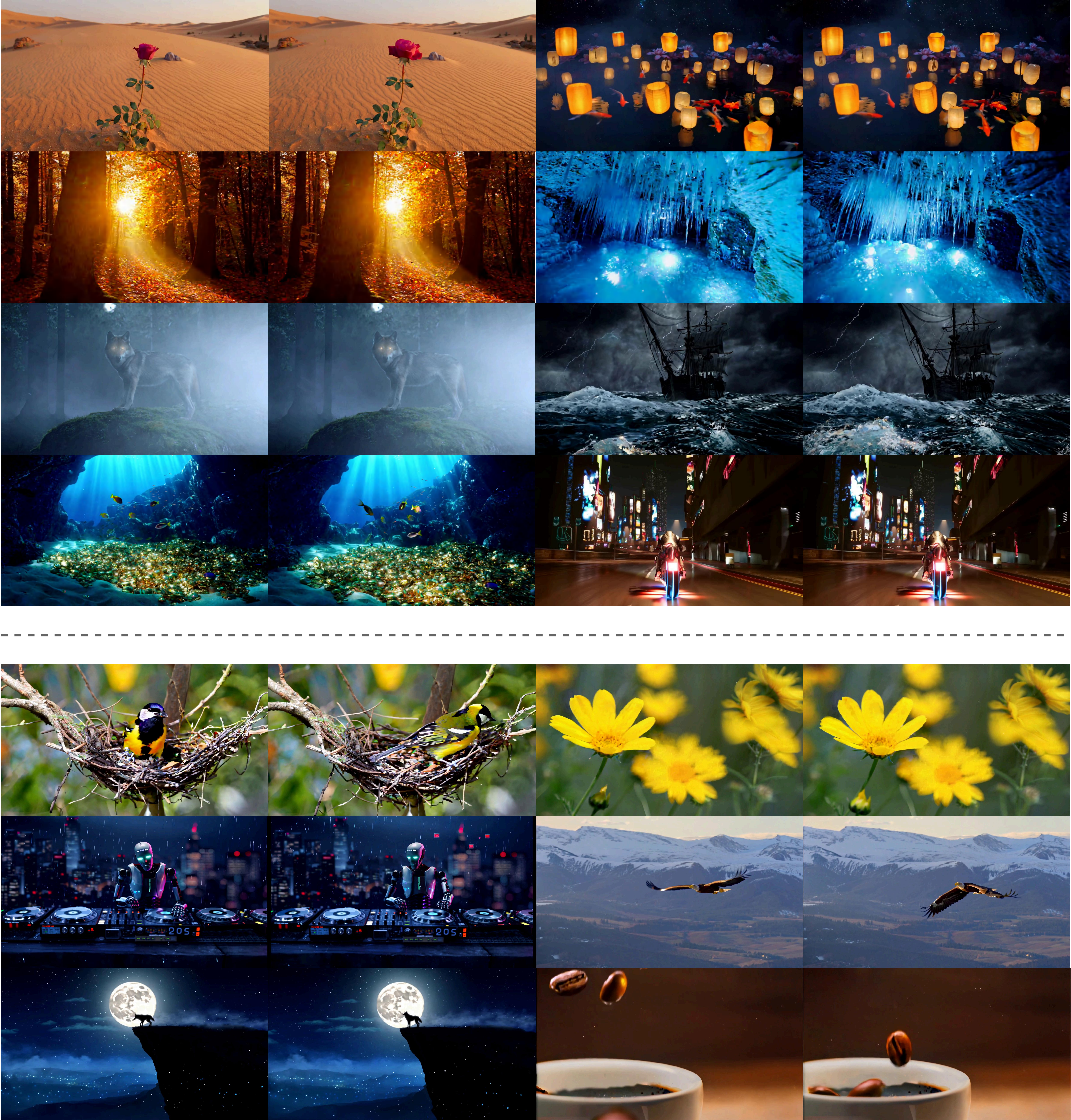}
    \caption{Qualitative results of 1080P generated by our method FreeSwim integrated into \textbf{Image-to-Video} and \textbf{Speech-to-Video} pipelines.}
    \label{fig:i2v s2v}
\end{figure*}

\begin{figure*}
    \centering
    \includegraphics[width=\linewidth]{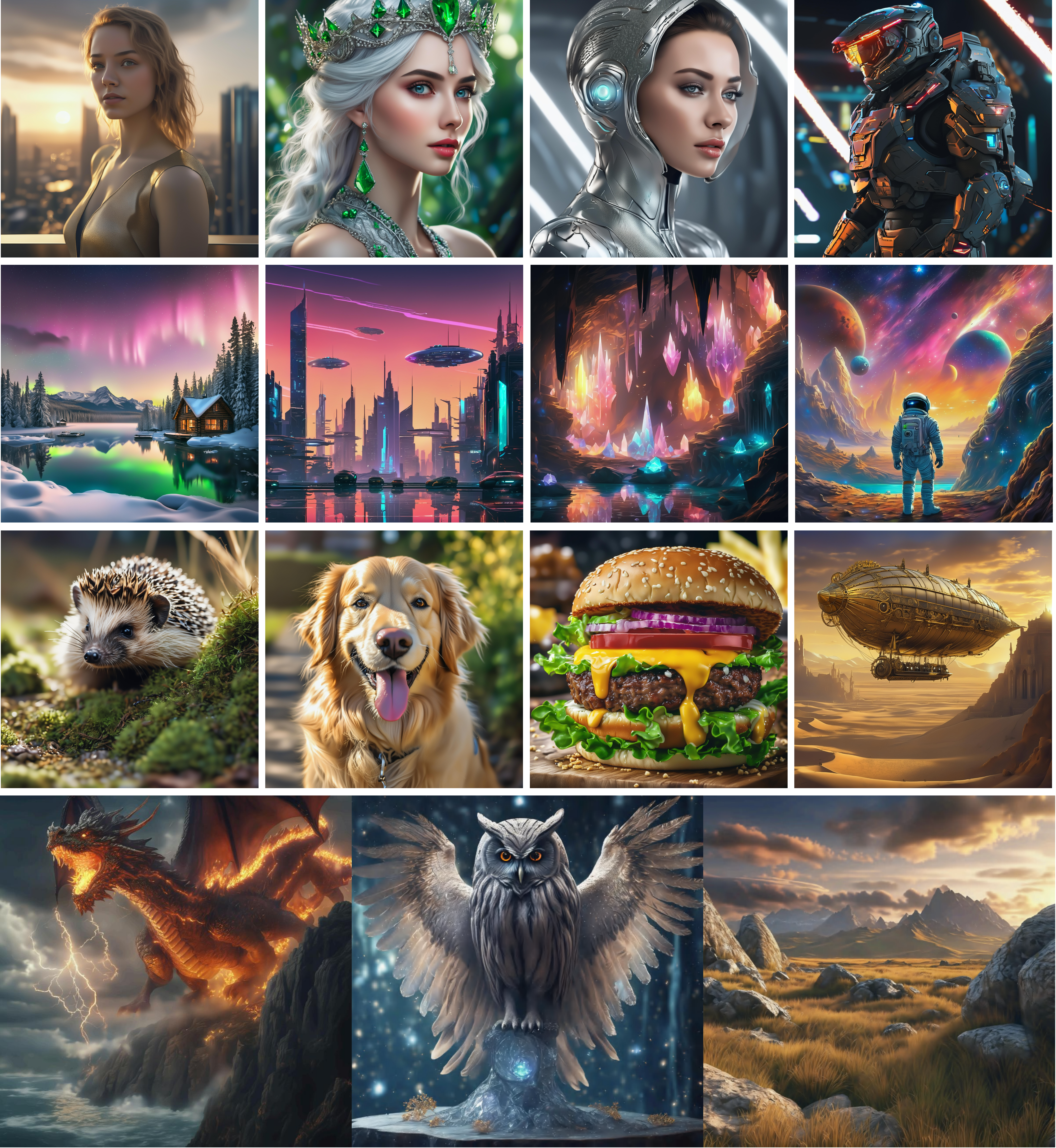}
    \caption{Qualitative results of 4K and 8K generated by our method FreeSwim integrated with SDXL~\cite{podell2023sdxlimprovinglatentdiffusion}.}
    \label{fig:sdxl}
\end{figure*}

\begin{figure*}
    \centering
    \includegraphics[width=\linewidth]{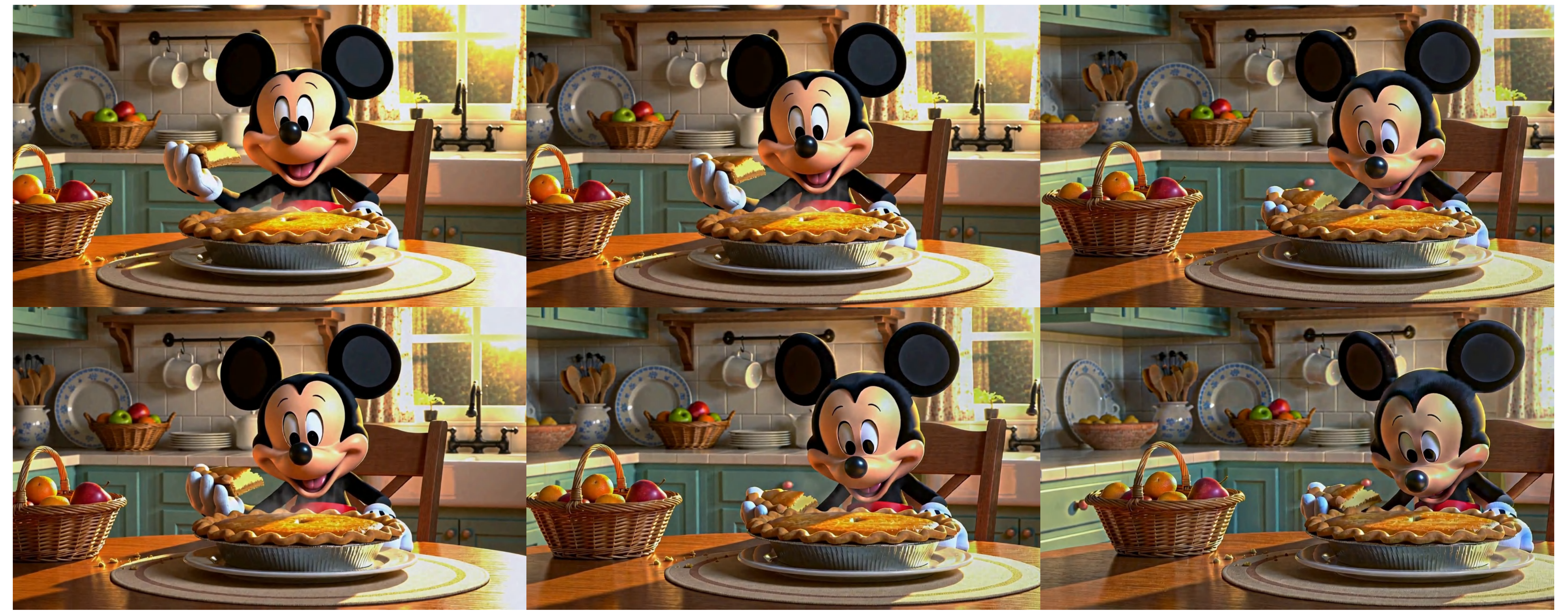}
    \caption{Qualitative results of $2\times$ upscaling in both height and width generated by our method FreeSwim integrated with Krea~\cite{krea_realtime_14b}}.
    \label{fig:krea}
\end{figure*}

\textbf{The breadth of our method:} 
As shown in Fig.~\ref{fig:sdxl}, our \textit{cross-attention override} can be seamlessly integrated into SDXL~\cite{podell2023sdxlimprovinglatentdiffusion}, enabling high-fidelity $4$K and $8$K image generation. Beyond still images, our method can be plugged into image-to-video, VACE, and speech-to-video pipelines, remaining training-free while leveraging the proposed coarse-to-fine strategy to generate high-resolution videos, as illustrated in Fig.~\ref{fig:vace} and~\ref{fig:i2v s2v}. For long videos, such as the $254$-frame sequence generated by Krea~\cite{krea_realtime_14b} under an autoregressive paradigm, the proposed \textit{inward sliding window attention} enables a $2\times$ upscaling in both height and width while still preserving fine visual details and global semantic structure, as shown in Fig.~\ref{fig:krea}.

\vspace{1.8cm}

\begin{tcolorbox}[title=Prompts for Video Generation,
                  colframe=black!20,
                  colback=white,
                  coltitle=black,
                  fonttitle=\bfseries,
                  breakable]
\footnotesize
\textbf{Figure 1 – Video grid.}\\
Images are arranged starting from the top-left corner, ordered row-wise from left to right and from top to bottom. 
The prompts are listed below in the same order. 
\begin{enumerate}[leftmargin=1.2em,itemsep=2pt]
  \item In the cold night sky, a mechanical eagle’s wings glimmer with cold light and fiery energy, reflecting the brilliant colors of the Aurora Borealis. The eagle flies over an ancient city covered in snow, its mechanical details and the light on its wings sharply defined in 8K resolution, creating a strong futuristic contrast against nature.

    \item A mermaid princess emerging from a vibrant coral landscape in an underwater realm, luminous silver hair cascading like waterfalls, singing a haunting melody, aerial perspective, blending fantasy and reality.

  \item Under the deep ocean, a ghostly ship floats between glowing jellyfish and spiral corals, its hull illuminated by glowing marine life. The blue fire from distant underwater volcanoes lights up the water, with every detail—from reflections to tiny glowing organisms—coming alive in 8K resolution, creating a dreamlike atmosphere with a haunting glow.

    \item A sleek black luxury sedan parked on a rain-soaked city street at night, reflecting neon lights from nearby buildings, smooth curves of the car highlighted by ambient urban glow.

  \item Surreal portrait of a young woman whose face merges with a complex arrangement of vivid flowers, bug’s-eye macro viewpoint from within petals emphasizing scale and perspective, vines and blossoms weaving seamlessly into translucent skin with freckles and delicate textures, diffused lighting casting soft shadows across blooms and contours, shallow depth of field with silky bokeh and fantastical elegance.

  \item A steampunk airship floating above a misty Victorian cityscape, intricate brass and copper mechanical details, golden hour lighting, detailed architectural elements, rich warm color palette.

  \item In a forest filled with metallic glimmers, a warrior in ancient armor rides a mechanized unicorn through a glowing jungle of electric plants and mechanical flowers. The sunlight pierces through heavy mist, casting brilliant rays on the copper and green mechanical structures. The scene is rich in detail, with a striking contrast between the future and nature, all rendered in stunning 8K clarity.

  \item A whimsical portrait of a fairy with luminescent wings, captured beneath a moonlit garden, flowing hair caught in an unseen breeze, intricate details on wings and petals in her hair, dreamy atmosphere.

  \item A detailed scene of an astronaut standing in a dense jungle, the astronaut's white space suit contrasting with the vibrant green foliage around. The jungle is alive with towering trees covered in thick vines, and exotic plants with bright colors peek through the shadows. Soft beams of light filter through the dense canopy above, casting intricate shadows on the astronaut’s suit and the surrounding flora. The astronaut’s visor reflects the jungle environment, adding depth to the scene. The overall color palette is cold and muted, with shades of green, blue, and gray dominating the image, creating a surreal contrast between the futuristic suit and the wild, untamed jungle. High-resolution 8k quality, highly detailed textures, and atmospheric lighting enhance the realism and intrigue of the scene.

  \item A sky bridge made of stardust and energy connects two floating planets in deep space. Beneath it, black holes and swirling nebulae create a mysterious backdrop. The glowing stars and shifting cosmic colors are rendered in breathtaking detail in 8K, with each light trail and star perfectly defined, creating a surreal and otherworldly visual experience.

  \item A colossal storm infused with mythical energy brews over an ancient landscape, where lightning arcs in radiant gold and violet across swirling clouds shaped like dragons and celestial beasts. Towering ruins of forgotten temples stand beneath the tempest, their stone carvings glowing with runic light as thunder tears through the heavens. The storm’s heart pulses with divine energy, merging nature’s fury and mystic power into a breathtaking cosmic spectacle, casting shimmering reflections on rain-soaked marble and turbulent seas.
\end{enumerate}

\vspace{6pt}
\textbf{Figure 2 – Video Examples.}
\begin{enumerate}[leftmargin=1.2em,itemsep=2pt]
  \item A realistic close-up of an elderly man with gray hair and a thick gray beard, wearing a light-colored shirt. His head is slightly lowered. The camera zooms from full body to close-up, highlighting detailed facial wrinkles, skin texture, forehead lines, eye bags, and beard strands. High resolution, cinematic lighting, sharp details.
\end{enumerate}

\vspace{6pt}

\textbf{Figure 5 – Case Study samples, also used in Figure 9.}
\begin{enumerate}[leftmargin=1.2em,itemsep=2pt]
  \item A majestic wolf standing proudly in a forest clearing, wearing intricately designed armor that gleams in the sunlight. The wolf moves slightly, its tail swishing gently and its ears flicking as it watches its surroundings. The armor is made of dark, polished metal with gold accents, and its design reflects the wolf's fierce and noble nature. The forest around the wolf is dense, with tall trees and soft beams of sunlight breaking through the canopy, casting shadows on the ground. The atmosphere is heroic and powerful, 8K.

  \item Enchanted forest with towering ancient trees and glowing bioluminescent plants, a mystical girl in flowing robes gently touching a glowing mushroom, magical aura surrounding her, fantasy style, vibrant yet natural colors, ethereal lighting, soft bokeh effect, ultra-detailed, painterly style, ultra HD, 8K, mist and fog weaving through the forest, otherworldly ambiance, soft glowing lights from flora only.

  \item A realistic close-up of an elderly man with silver-gray hair and a soft beard, sitting on sunlit grass. He wears a light linen shirt and a beige vest, his face illuminated by warm afternoon sunlight filtering through the trees. The camera moves from a medium shot to a close-up, capturing defined wrinkles, sun-kissed skin texture, and individual beard strands. His eyes sparkle with vitality, exuding a sense of wisdom and energy. The background is softly blurred with golden bokeh, highlighting the lively and confident expression on his face. High realism, natural lighting, ultra-detailed texture, 8K resolution, cinematic depth of field.
\end{enumerate}

\vspace{6pt}

\textbf{Figure 10 – Case Study.}

\begin{enumerate}[leftmargin=1.2em,itemsep=2pt]
  \item A warm and atmospheric scene inside an ancient library illuminated only by flickering candlelight. Towering bookshelves filled with weathered tomes stretch into the shadows, while dust particles drift gently in the golden glow. A single wooden desk sits in the center, its polished surface reflecting the soft light from a tall candle, casting long, dancing shadows on the surrounding stone walls. The rich textures of leather-bound books, carved wood, and aged parchment create a sense of timeless wisdom. High-resolution 8k quality, ultra-detailed textures, and warm, natural lighting emphasize the quiet, scholarly mood of the setting.

  \item A gentle riverside scene where cherry blossom petals drift softly across the water in the spring breeze. The branches above are full of pink blossoms, some catching warm sunlight, others swaying gently in the wind. The river’s surface ripples delicately, reflecting the soft pastel sky. Smooth stones along the bank are partially covered by fallen petals, adding to the tranquil atmosphere. High-resolution 8k quality, detailed natural textures, and soft warm lighting emphasize the peaceful and poetic beauty of the scene.

    \item A sweeping view of a desert caravan at sunset, with camels walking slowly across golden dunes painted by the warm glow of the dying sun. Silhouettes of distant travelers appear against the horizon, their robes catching subtle highlights from the orange sky. Soft sands ripple underfoot, with intricate textures shaped by the wind. The sky transitions from fiery gold to soft purples, with thin clouds drifting lazily overhead. High-resolution 8k quality, dynamic lighting, and ultra-detailed textures capture the elegance and quiet determination of the desert travelers.

  \item A detailed scene of a lone wolf standing on a rocky cliff at dawn, gazing out over a vast, misty valley below. The wolf's fur glistens in the soft, golden light of the rising sun, and the surrounding rocks are textured and weathered, with subtle moss growing in the crevices. The sky above is painted in soft pinks and oranges, with a few wispy clouds drifting by. The valley below is covered in fog, with a faint silhouette of distant mountains barely visible. The background is simple yet striking, with a focus on the wolf’s powerful stance and the warm light highlighting its fur. High-resolution 8k quality, ultra-detailed textures, and dynamic lighting emphasize the serenity and power of the wolf.

    \item A serene close-up of a brightly colored hummingbird hovering near a simple flower, its iridescent feathers shimmering in the sunlight. The background is a smooth gradient of soft blues and greens, with a slight bokeh effect to keep the focus on the bird. The flower, with its delicate petals, stands still in the gentle breeze, with dewdrops glistening on the petals and the bird’s wings. The composition highlights the beauty of nature in a minimalist setting. High-resolution 8k quality, highly detailed textures, and soft, natural lighting enhance the beauty and grace of the scene.

  \item A magical twilight meadow filled with glowing fireflies drifting among tall grasses. The sky transitions from deep blue to soft purple as the last light of day fades, and the fireflies create a constellation-like pattern across the field. Wildflowers sway gently in the cool breeze, their delicate petals catching hints of moonlight. In the distance, a silhouette of a lone tree adds depth to the tranquil scene. High-resolution 8k quality, ultra-detailed natural elements, and soft atmospheric lighting enhance the enchanted, peaceful mood.

  \item A snow-covered mountain peak under a brilliant starry sky, with the Milky Way clearly visible in the background. The snow is crisp and pure, reflecting the pale light of the moon, while sharp ridges and rocky outcrops create a dramatic contrast against the smooth snow. A soft wind blows, causing a slight drift of snow, creating a dynamic and serene atmosphere. The background is a deep black sky filled with stars, the Milky Way stretching across, adding a sense of vastness and wonder. High-resolution 8k quality, ultra-detailed textures, and natural lighting capture the breathtaking beauty of this remote, pristine landscape.

  \item A peaceful forest scene at dawn, where a clear creek winds gently between moss-covered rocks. Soft morning mist hangs low among the tall trees, and beams of golden sunlight filter through the dense canopy, creating shimmering reflections on the water’s surface. Delicate ferns line the banks, their dew-covered leaves glistening in the early light. The background remains softly blurred to highlight the quiet beauty of the creek and its surrounding greenery. High-resolution 8k quality, extremely detailed textures, and soft natural lighting create a serene, immersive atmosphere.
  
  \item A fiery phoenix soaring through a golden sky, its wings glowing with intense flames. The bird’s feathers are detailed with fiery orange and red hues, and its body leaves a trail of embers in the air. The sun sets in the background, casting warm light across the scene. High-resolution 8k quality, ultra-detailed textures, and dramatic lighting emphasize the power and grace of this mythical creature.

  \item A majestic lion standing on a golden savannah at sunset, the warm glow of the setting sun casting long shadows across the grass. The lion’s mane is thick and flowing, bathed in golden light, with each individual strand detailed against the backdrop of a fading sky painted in oranges, reds, and purples. The grass sways gently in the breeze, and the distant hills are softly blurred, providing a sense of depth. The scene captures the power and elegance of the lion in its natural habitat. High-resolution 8k quality, ultra-detailed textures, and dynamic lighting emphasize the serene yet commanding presence of the lion.

    \item A glowing jellyfish floating through the dark ocean depths, its bioluminescent tentacles trailing behind it. The soft, otherworldly light from the jellyfish illuminates the surrounding water, revealing a distant underwater landscape of smooth rocks and drifting seaweed. High-resolution 8k quality, ultra-detailed textures, and natural lighting capture the serene, mysterious beauty of the deep ocean.

  \item A sleek black motorcycle parked on a quiet, rain-drenched street at night, its chrome parts reflecting the neon glow from nearby signs. The road is wet, with puddles reflecting the vibrant lights of the city around it. The motorcycle’s design is sharp and modern, with every curve and detail sharply defined against the glossy street. In the background, the soft blur of city lights creates a moody and cinematic atmosphere. High-resolution 8k quality, sharp details, and atmospheric lighting enhance the futuristic yet grounded vibe of the scene.

  \item A powerful scene of waves crashing against a rugged ocean cliff during a brewing thunderstorm. Dark clouds churn overhead, illuminated occasionally by bright flashes of lightning that reveal the textures of the cliffside. Seaspray rises with each impact of the waves, catching subtle highlights from the stormy sky. Wind-bent grasses cling to the rocks, adding movement and life to the scene. High-resolution 8k quality, dramatic lighting, and ultra-detailed natural textures emphasize the raw power of nature.

  \item A close-up of a beautifully crafted wooden violin resting on a soft velvet cloth, surrounded by delicate rose petals. The violin’s rich, polished surface gleams in the warm light, and the intricate curves and fine details of the wood are visible, reflecting the surrounding light. The cloth beneath is textured, and the soft rose petals create a romantic and elegant composition. The background is a smooth blur of warm tones, allowing the focus to stay on the violin and its refined details. High-resolution 8k quality, highly detailed textures, and warm, soft lighting enhance the beauty and craftsmanship of the scene.

  \item A calm scene of a cat sitting on a windowsill, gazing out at a rainy street. The soft raindrops create delicate streaks on the window, while the cat's fur is sleek and shiny under the warm glow of a nearby lamp. The background is a simple, darkened cityscape with blurred streetlights and distant buildings. The focus is on the peaceful moment, with the subtle contrast between the warm, cozy interior and the cool, rainy world outside. High-resolution 8k quality, soft textures, and atmospheric lighting create an intimate, peaceful atmosphere.

  \item A nostalgic scene of a vintage train station at dawn, with a steam locomotive releasing gentle plumes of white smoke into the cool morning air. Warm orange sunlight begins to spill across the platform, illuminating the glossy metal of the train and the textured wooden beams of the station roof. A few early passengers stand quietly, their silhouettes softened by the drifting steam. The rails glisten with dew as the light touches them. High-resolution 8k quality, ultra-detailed surfaces, and soft cinematic lighting capture the peaceful anticipation of early travel.

  \item A lone astronaut standing on the edge of a crater on Mars, with the planet’s surface stretching into the distance. The orange-red hues of the Martian landscape contrast with the bright white of the astronaut's spacesuit. In the background, distant mountains rise against the soft, hazy sky, while the faint silhouette of a rover can be seen parked near a rock formation. High-resolution 8k quality, ultra-detailed textures, and dramatic lighting create a cinematic, solitary feel to the scene.

  \item A serene close-up of a glowing firefly resting on a delicate leaf in a dark forest. The soft, greenish glow of the firefly illuminates the surrounding dark foliage, casting a gentle light on the leaf. The background is a smooth blur of dark trees and soft bokeh, creating a peaceful, magical atmosphere. High-resolution 8k quality, ultra-detailed textures, and soft lighting capture the tranquil, enchanting beauty of this tiny creature.

\end{enumerate}

\vspace{6pt}

\textbf{Figure 11 – VACE samples.}

\begin{enumerate}[leftmargin=1.2em,itemsep=2pt]
  \item An antique film camera resting on a wooden desk, with soft light casting gentle shadows across the polished surface. The camera’s metal parts and intricate design are highlighted by the warm glow, while the leather casing shows subtle signs of age. A few vintage film reels sit next to the camera, their metallic spools glinting in the light. The background features a blurred bookshelf filled with old photography books, adding depth and character to the scene. High-resolution 8k quality, ultra-detailed textures, and natural lighting emphasize the nostalgic and elegant vibe of the scene.

  \item An antique pocket watch lying on an old velvet cloth, its intricate design and gold casing gleaming in the soft light. The timepiece’s hands are frozen in place, adding a sense of mystery. The delicate chain rests beside it, coiled in an elegant loop. The background is softly blurred, focusing on the fine details of the watch’s surface, the worn velvet, and the soft play of light. High-resolution 8k quality, ultra-detailed textures, and soft lighting evoke a sense of history and timelessness.

    \item A sleek black motorcycle parked on a quiet, rain-drenched street at night, its chrome parts reflecting the neon glow from nearby signs. The road is wet, with puddles reflecting the vibrant lights of the city around it. The motorcycle’s design is sharp and modern, with every curve and detail sharply defined against the glossy street. In the background, the soft blur of city lights creates a moody and cinematic atmosphere. High-resolution 8k quality, sharp details, and atmospheric lighting enhance the futuristic yet grounded vibe of the scene.

  \item A close-up of a beautifully crafted wooden violin resting on a soft velvet cloth, surrounded by delicate rose petals. The violin’s rich, polished surface gleams in the warm light, and the intricate curves and fine details of the wood are visible, reflecting the surrounding light. The cloth beneath is textured, and the soft rose petals create a romantic and elegant composition. High-resolution 8k quality, highly detailed textures, and warm, soft lighting enhance the beauty and craftsmanship of the scene..

    \item A serene close-up of a brightly colored hummingbird hovering near a simple flower, its iridescent feathers shimmering in the sunlight. The background is a smooth gradient of soft blues and greens, with a slight bokeh effect to keep the focus on the bird. The flower, with its delicate petals, stands still in the gentle breeze, with dewdrops glistening on the petals and the bird’s wings. The composition highlights the beauty of nature in a minimalist setting. High-resolution 8k quality, highly detailed textures, and soft, natural lighting enhance the beauty and grace of the scene.

  \item An elegant vintage ink pen resting on a piece of aged parchment, with the ink still fresh and glistening on the paper. The pen’s intricate design and ornate nib are highlighted by soft, warm lighting. The paper is textured, and the faint writing visible on it adds to the vintage feel. In the background, an antique ink bottle and a few rolled-up letters add depth to the composition. High-resolution 8k quality, ultra-detailed textures, and soft lighting enhance the old-world charm of the scene.

  \item A fresh assortment of fruits arranged neatly on a delicate porcelain plate, with each fruit’s vibrant color shining under soft natural light. The fruits are meticulously arranged—plump grapes, ripe strawberries, and bright citrus fruits add vivid color to the scene. The porcelain plate’s intricate design is visible, and a few scattered mint leaves provide a fresh touch. The background is softly blurred to keep the focus on the fruits. High-resolution 8k quality, ultra-detailed textures, and soft natural lighting enhance the fresh and inviting atmosphere..

  \item A freshly baked loaf of bread resting on a rustic wooden cutting board, with the warm golden-brown crust glistening slightly under soft lighting. The bread’s texture is visible, with a few crumbs scattered around the board. A sharp bread knife rests beside the loaf, adding to the rustic charm. The background is softly blurred to highlight the bread’s texture and the warmth of the scene. High-resolution 8k quality, ultra-detailed textures, and soft lighting create a comforting, homely atmosphere.
  
  \item A gentle riverside scene where cherry blossom petals drift softly across the water in the spring breeze. The branches above are full of pink blossoms, some catching warm sunlight, others swaying gently in the wind. The river’s surface ripples delicately, reflecting the soft pastel sky. Smooth stones along the bank are partially covered by fallen petals, adding to the tranquil atmosphere. High-resolution 8k quality, detailed natural textures, and soft warm lighting emphasize the peaceful and poetic beauty of the scene.

  \item An ancient spellbook lies open on a stone pedestal, its pages glowing faintly with floating golden runes that swirl gently in the air. The book’s cover is embossed with arcane symbols, and faint embers drift upward like fireflies. Surrounding candles flicker softly, casting warm light on the worn leather and metallic clasps. The atmosphere feels mystical and cinematic, filled with wonder and mystery. High-resolution 8k quality, ultra-detailed textures, and soft glowing light create a fantasy-inspired visual masterpiece.

    \item A set of crystal glassware, elegantly arranged on a wooden table, with soft lighting highlighting the delicate cuts and reflections on the glass. The glasses are filled with sparkling water, catching the light and creating beautiful reflections. The background is softly blurred, focusing on the intricate details of the glassware and the warm tones of the wooden table. High-resolution 8k quality, ultra-detailed textures, and natural lighting emphasize the elegance and clarity of the glassware..

  \item A sleek futuristic desk with a glowing blue holographic interface projected above its surface. The transparent digital panels display dynamic graphs and 3D data models, reflecting light onto the metallic desk edges. A glass cup with condensation sits nearby, grounding the high-tech environment. The background shows a panoramic city skyline filled with neon lights through a large glass window. High-resolution 8k quality, ultra-detailed textures, and cinematic lighting emphasize the advanced sci-fi atmosphere.

  \item An artistic workspace featuring a wooden easel, colorful paintbrushes, and a palette filled with vivid pigments. The table is covered in splashes of paint, capturing the energy of creativity. A half-finished painting sits on the easel, showing a vibrant landscape under soft daylight. The brushes and tubes of paint are scattered naturally, while the soft texture of the canvas and wood grains add realism. High-resolution 8k quality, ultra-detailed textures, and balanced natural lighting evoke the warmth of an artist’s studio.

  \item A cozy morning scene showing a steaming cup of coffee placed beside a folded newspaper on a wooden breakfast table. Sunlight streams through a nearby window, casting warm golden tones across the table surface. Wisps of steam curl above the coffee cup, and the reflections dance on the ceramic glaze. A small silver spoon rests nearby, and soft shadows create a sense of tranquility. High-resolution 8k quality, ultra-detailed textures, and warm natural lighting emphasize the comfort and calm of a peaceful morning.

  \item A fluffy cat sleeping peacefully beside a softly flickering candle on a small wooden table. The candle’s warm glow illuminates the cat’s fur, revealing its gentle texture and casting a soft halo of light. The background is dark and cozy, with faint reflections on nearby glass surfaces. A closed book and a folded blanket complete the quiet, intimate atmosphere. High-resolution 8k quality, ultra-detailed textures, and warm cinematic lighting emphasize the tranquility and warmth of the moment.

  \item A single vintage street lantern glowing warmly in a narrow cobblestone alley during a gentle rain. Raindrops glisten on the stones, reflecting the amber light, while mist rises from the ground, adding a moody softness to the scene. Old brick walls and a few hanging signs fade into the background blur, creating a timeless urban charm. High-resolution 8k quality, ultra-detailed textures, and cinematic lighting emphasize the atmospheric depth and tranquility of the rainy night.

  \item A sleek, black sports car parked on a foggy bridge at dawn, with city lights twinkling in the background. The car's curves are highlighted by the soft glow of the early morning light, and the mist rolls over the bridge, partially obscuring the distant skyline. The sleek design of the car contrasts against the soft, ethereal fog, creating a striking visual balance. High-resolution 8k quality, ultra-detailed textures, and atmospheric lighting enhance the futuristic, cool vibe of the scene.

  \item A large tortoise with a crystalline, gem-like shell resting beside a sparkling riverbank at dawn. Early sunlight refracts through the translucent shell, scattering tiny rainbow glimmers onto the moss-covered rocks. Water flows slowly in the background, reflecting the soft pink morning sky. Dragonflies with metallic wings hover nearby, adding gentle movement to the tranquil scene. High-resolution 8k quality, ultra-detailed textures, natural lighting combined with fantasy elements create a serene, magical realism style..

\end{enumerate}

\vspace{6pt}

\textbf{Figure 12 – Image-to-Video and Speech-to-Video samples.}
\begin{enumerate}[leftmargin=1.2em,itemsep=2pt]
  \item A single desert rose blooming in the middle of a vast, dry desert. The contrast between the delicate, vibrant petals and the harsh, arid landscape creates a striking visual. The rose is bathed in soft sunlight, with sand dunes stretching out into the distance. A few scattered rocks and cacti provide minimal contrast in the surrounding barren environment. The scene captures the beauty of resilience in nature. High-resolution 8k quality, ultra-detailed textures, and natural lighting emphasize the contrast between fragility and strength in the desert landscape.

    \item A tranquil koi pond under a starlit night, with glowing lanterns floating gently on the water’s surface. The koi fish glide gracefully beneath the reflections, their scales shimmering in shades of orange, white, and gold. Ripples distort the reflections of the paper lanterns, while soft moonlight illuminates nearby lotus flowers. Mist hovers above the pond, creating an ethereal, almost magical atmosphere. High-resolution 8k quality, ultra-detailed textures, and soft, diffused lighting capture the quiet beauty of this serene night scene.

  \item A sunlit autumn forest filled with golden and crimson leaves, gently falling through beams of warm afternoon light. A narrow path winds through the trees, covered in a soft carpet of fallen leaves. Shafts of sunlight pierce the canopy, illuminating particles of dust and creating a glowing, dreamlike effect. The bark of the trees is richly textured, and the atmosphere feels calm and nostalgic. High-resolution 8k quality, ultra-detailed textures, and soft golden lighting emphasize the warmth and serenity of the autumn forest.

  \item A breathtaking ice cavern with crystal-clear ice formations hanging from the ceiling, reflecting the pale light that filters through narrow openings. The cavern’s walls shimmer with frost, and the icy stalactites and stalagmites create a dazzling display of natural art. Soft blue light from glowing ice crystals provides the only illumination, casting an ethereal glow across the cavern floor. High-resolution 8k quality, ultra-detailed textures, and soft, ambient lighting emphasize the serene yet cold beauty of this frozen world.

    \item A lone wolf stands on a moss-covered rock under a full moon, its fur shimmering with silvery highlights as it gazes into the distance. The surrounding forest is dark, with shadows cast by tall trees and soft light filtering through the leaves. The wolf's piercing eyes glow faintly, creating an eerie yet majestic presence. Mist rises from the forest floor, and distant howls can be heard in the background. High-resolution 8k quality, ultra-detailed textures, and soft lighting emphasize the mystical and wild atmosphere of the forest.

  \item A dramatic scene of a pirate ship battling the furious waves of a stormy sea. Dark, swirling clouds fill the sky, and bolts of lightning strike the water, creating intense reflections on the surface. The pirate ship's sails are torn, and the crew can be seen battling the storm on deck. The ship creaks under the force of the waves, adding a sense of urgency and chaos. High-resolution 8k quality, ultra-detailed textures, and dynamic lighting enhance the intensity and drama of the scene.

  \item A hidden underwater cave filled with glittering treasures—golden coins, sparkling jewels, and ancient artifacts. Soft rays of sunlight penetrate the clear blue water, illuminating the treasures scattered across the sandy seabed. Schools of colorful fish swim through the scene, and bioluminescent plants glow softly on the cave walls. The atmosphere is serene yet mysterious, with a sense of discovery. High-resolution 8k quality, ultra-detailed textures, and natural lighting capture the magical allure of this underwater treasure cove.

  \item A futuristic motorcycle speeding down a neon-lit city street at night. The motorcycle’s sleek body glows with pulsing lights as it zooms past glowing billboards and towering skyscrapers. The streets are wet from rain, with reflections of neon signs shimmering on the asphalt. High-resolution 8k quality, sharp details, and dynamic lighting enhance the high-energy, cyberpunk aesthetic of the scene.
  
  \item A bird building a nest from twigs and leaves

  \item Yellow flowers swing in the wind

  \item A robot DJ is playing the turntable, in heavy raining futuristic tokyo rooftop cyberpunk night, sci-fi, fantasy

  \item A majestic eagle soaring high above snow-capped mountains, its wings spread wide as it catches the wind. The eagle’s feathers are sharp and detailed, and its piercing eyes are focused ahead. The landscape below is vast and untouched, with valleys, forests, and mountain peaks stretching into the distance. The sky is clear, with soft, golden light highlighting the eagle and its surroundings. High-resolution 8k quality, ultra-detailed textures, and dynamic lighting capture the power and grace of this wild bird in flight.

    \item A lone wolf standing on the edge of a moonlit cliff, its silhouette glowing under a bright full moon. Soft mist drifts around the cliffside, and the wolf’s fur ripples gently in the night breeze. Below, a vast valley stretches into darkness, dotted with faint blue lights resembling fireflies. The sky is filled with stars, and thin clouds move slowly across the moon, creating a dramatic, cinematic atmosphere. High-resolution 8k quality, ultra-detailed textures, and cool-toned lighting highlight the mystery and strength of the wolf.

  \item Macro slo-mo. Slow motion cropped closeup of roasted coffee beans falling into an empty bowl.
  
\end{enumerate}

\vspace{6pt}

\textbf{Figure 13 – Image samples.}
\begin{enumerate}[leftmargin=1.2em,itemsep=2pt]
  \item A photorealistic portrait of a young woman standing on a futuristic city balcony at sunset. Soft golden sunlight illuminates her face, highlighting natural skin texture and fine hair strands. Behind her, ultra-detailed skyscrapers with glass reflections, flying vehicles, and gentle atmospheric haze. Realistic lighting, shallow depth of field, natural color grading, 8K, extremely lifelike details, cinematic realism.

    \item Ultra HD 8K resolution, hyper-realistic, insanely detailed, sharp focus, Moonlight elf princess in deep forest, silver hair, emerald eyes, diamond vine crown, dappled light, pore-level skin detail, shot on Canon 85 mm f/1.4

  \item Elegant futuristic female android with silver metallic skin, soft expression, volumetric light, highly realistic skin texture, sci-fi portrait, 8K.

  \item Futuristic soldier, robotic armor, high-tech weapon, visor with digital HUD, dark sci-fi, highly detailed, cinematic lighting, dynamic pose, ultra-realistic, ultra HD, 8K, neon accents, dark background, glowing HUD, intense expression, battle scars on armor.

    \item A tranquil alpine lake reflecting the Northern Lights, wooden cabin smoke rising, snow-covered pines, cozy winter vibe.

  \item Cyberpunk city street at night, neon reflections on wet ground, flying cars, holograms, dense atmosphere, cinematic lighting, ultra-detailed, dramatic perspective.

  \item Glowing crystals in a dark cave, soft ambient light, intricate reflections, ultra-detailed fantasy environment.

  \item Astronaut meeting alien creatures, cosmic background, colorful nebula, stars in background, high detail spacesuit, atmospheric lighting, sci-fi setting, calm and peaceful, otherworldly creatures, ultra-realistic, adventure in space, detailed environment.

  \item Realistic hedgehog exploring mossy ground, shallow depth of field, macro lens effect, highly detailed quills and textures.

  \item German shepherd running across a field, motion-blurred background, crisp focus on the dog, dynamic action shot, 4K sports-style photography.

  \item Ultra HD 8K resolution, hyper-realistic, insanely detailed, sharp focus, Fresh Angus beef burger, molten cheese, water-droplet veggies, toasted sesame bun, macro food shot, f/2.8, mouth-watering

    \item Steampunk airship floating above a misty Victorian cityscape, intricate brass and copper mechanical details, golden hour lighting, billowing clouds, detailed architectural elements, rich warm color palette, cinematic composition.

  \item An elemental sorcerer summoning flames atop a cliff, a colossal dragon soaring through storm clouds, lightning illuminating the scene, dramatic composition, epic fantasy illustration, powerful contrast, 8K exquisite details.

  \item Giant owl guardian perched on a crystal tree, magical particles drifting, dark fantasy mood.

  \item Ultra HD 8K panoramic mountain landscape, golden hour light, hyper-realistic rocks and grass textures, dramatic sky with volumetric clouds.
\end{enumerate}

\vspace{6pt}

\vspace{6pt}

\textbf{Figure 14 – Long Video samples.}
\begin{enumerate}[leftmargin=1.2em,itemsep=2pt]
  \item A cheerful Mickey Mouse sits at a cozy wooden table in a warmly lit kitchen, happily eating a freshly baked pie. The golden crust of the pie glistens under the soft light, with steam gently rising from the filling. Mickey’s eyes sparkle with delight as he takes a big bite, crumbs scattering playfully on the tablecloth. Behind him, the kitchen is filled with soft pastel tones, with a few plates, fruit baskets, and a window showing the golden glow of an afternoon sun. High-resolution 8k quality, ultra-detailed textures, and warm, cinematic lighting create a charming and nostalgic atmosphere.

\end{enumerate}

\textbf{Figure 15 – HunYuanVideo samples.}
\begin{enumerate}[leftmargin=1.2em,itemsep=2pt]
  \item A realistic close-up of an elderly man with silver-gray hair and a soft beard, sitting on sunlit grass. He wears a light linen shirt and a beige vest, his face illuminated by warm afternoon sunlight filtering through the trees. The camera moves from a medium shot to a close-up, capturing defined wrinkles, sun-kissed skin texture, and individual beard strands. His eyes sparkle with vitality, exuding a sense of wisdom and energy. The background is softly blurred with golden bokeh, highlighting the lively and confident expression on his face. High realism, natural lighting, ultra-detailed texture, 8K resolution, cinematic depth of field

    \item  A detailed scene of an astronaut standing in a dense jungle, the astronaut's white space suit contrasting with the vibrant green foliage around. The jungle is alive with towering trees covered in thick vines, and exotic plants with bright colors peek through the shadows. Soft beams of light filter through the dense canopy above, casting intricate shadows on the astronaut’s suit and the surrounding flora. The astronaut’s visor reflects the jungle environment, adding depth to the scene. The overall color palette is cold and muted, with shades of green, blue, and gray dominating the image, creating a surreal contrast between the futuristic suit and the wild, untamed jungle. High-resolution 8k quality, highly detailed textures, and atmospheric lighting enhance the realism and intrigue of the scene.
    
    \item A majestic wolf standing proudly in a forest clearing, wearing intricately designed armor that gleams in the sunlight. The wolf moves slightly, its tail swishing gently and its ears flicking as it watches its surroundings.

    \item Snow-covered mountains under a starry sky, timelapse of aurora borealis dancing across the horizon.

    \item A space station orbiting Earth, astronauts floating, the planet slowly rotating below.

    \item A serene close-up of a glowing firefly resting on a delicate leaf in a dark forest. The soft, greenish glow of the firefly illuminates the surrounding dark foliage, casting a gentle light on the leaf. The background is a smooth blur of dark trees and soft bokeh, creating a peaceful, magical atmosphere. High-resolution 8k quality, ultra-detailed textures, and soft lighting capture the tranquil, enchanting beauty of this tiny creature.

\end{enumerate}

\end{tcolorbox}

\begin{tcolorbox}[
    title=Hyperparameters for Video Generation Experiments,
    colframe=black!20,
    colback=white,
    coltitle=black,
    fonttitle=\bfseries,
    breakable
]
\begin{itemize}
    \item \texttt{num\_inference\_steps} = 50
    \item \texttt{strength} = 0.7
    \item \texttt{guidance\_scale} = 5.0
    \item \texttt{max\_sequence\_length} = 512
    \item \texttt{flow\_shift} = 9.0
    \item \texttt{attention\_scale = math.sqrt(math.log(30 $\times$ 52 $\times$ 2, 30 $\times$ 52) / key.size(3))}
\end{itemize}
\end{tcolorbox}


\begin{tcolorbox}[
    title=Evaluation by VLM,
    colframe=black!20,
    colback=white,
    coltitle=black,
    fonttitle=\bfseries,
    breakable
]
\begin{table}[H]
\centering
\setlength{\tabcolsep}{1.1pt}
\resizebox{\linewidth}{!}{
\renewcommand{\arraystretch}{1.18}
\begin{tabular}{l|ccc}
\toprule
\textbf{Model} &
\makecell{Text\\Alignment} &
\makecell{Aesthetic\\Quality} &
\makecell{Video\\Quality} \\
\midrule
\textbf{Real-ESRGAN~\cite{wang2021realesrgantrainingrealworldblind}} & 9.0 & 8.2 & 8.5 \\
\textbf{UAV~\cite{zhou2023upscaleavideotemporalconsistentdiffusionmodel}} & 8.7 & 8.3 & 8.2 \\
\textbf{I-Max~\cite{du2024imaxmaximizeresolutionpotential}} & 8.7 & 8.6 & 8.3 \\
\textbf{HiFLow~\cite{bu2025hiflowtrainingfreehighresolutionimage}} & 8.1 & 8.3 & 7.8 \\
\textbf{Direct Inference~\cite{wan2025wanopenadvancedlargescale}} & 7.8 & 7.8 & 8.1 \\
\textbf{CineScale \textit{w/o} LoRA~\cite{qiu2025cinescalefreelunchhighresolution}} & 6.5 & 7.0 & 7.0 \\
\textbf{CineScale \textit{w/} LoRA~\cite{qiu2025cinescalefreelunchhighresolution}} & 7.8 & 8.4 & 8.2 \\
\textbf{Ours} & 9.3 & 9.2 & 8.8 \\
\bottomrule
\end{tabular}
}
\caption{Model Performance Averages for Text Alignment, Aesthetic Quality, and Video Quality evaluated by GPT-5.}
\label{tab:model performance}
\end{table}
\end{tcolorbox}

\newpage







\end{document}